\documentclass[final,3p,times]{elsarticle}

\usepackage{amsmath}
\usepackage{amssymb}
\usepackage{graphicx}
\usepackage{caption}
\usepackage{float}
\usepackage{subfigure}
\usepackage{tabularx}
\usepackage{multirow}
\usepackage[ruled,linesnumbered]{algorithm2e}  
\usepackage[noend]{algpseudocode}

\usepackage[figuresright]{rotating}

\begin{document}

\begin{frontmatter}



\title{A novel time--frequency Transformer based on self--attention mechanism and its application in fault diagnosis of rolling bearings}


\author[]{Yifei Ding}
%\ead{ifei.ting@outlook.com}

\author[]{Minping Jia\corref{cor1}}
\ead{mpjia@seu.edu.cn}

\author[]{Qiuhua Miao}

\author[]{Yudong Cao}

\address[]{School of Mechanical Engineering, Southeast University, Nanjing 211189, PR China}

\cortext[cor1]{Corresponding author}

\begin{abstract}
 The scope of data-driven fault diagnosis models is greatly extended through deep learning (DL). However, the classical convolution and recurrent structure have their defects in computational efficiency and feature representation, while the latest Transformer architecture based on attention mechanism has not yet been applied in this field. To solve these problems, we propose a novel time--frequency Transformer (TFT) model inspired by the massive success of vanilla Transformer in sequence processing. Specially, we design a fresh tokenizer and encoder module to extract effective abstractions from the time--frequency representation (TFR) of vibration signals. On this basis, a new end-to-end fault diagnosis framework based on time--frequency Transformer is presented in this paper. Through the case studies on bearing experimental datasets, we construct the optimal Transformer structure and verify its fault diagnosis performance. The superiority of the proposed method is demonstrated in comparison with the benchmark models and other state-of-the-art methods.
\end{abstract}


%	\item A novel model named time--frequency Transformer (TFT) is proposed.
%	\item A fresh tokenizer and encoder module are designed to extract effective abstractions.
%	\item A new end-to-end fault diagnosis framework based on TFT is presented.
%	\item The proposed method shows superiority comparing to other state-of-the-art methods.

\begin{keyword}
	Fault diagnosis \sep Deep learning \sep Transformer \sep Self--attention mechanism \sep Rolling bearings
\end{keyword}

\end{frontmatter}


\section{Introduction}
\label{sec:introduction}
%\begin{itemize} \item document style \item baselineskip \item front
%matter \item keywords and MSC codes \item theorems, definitions and
%proofs \item lables of enumerations \item citation style and labeling.
%\end{itemize}
%
%This class depends on the following packages
%for its proper functioning:
%
%\begin{enumerate}
%\itemsep=0pt
%\item {natbib.sty} for citation processing;
%\item {geometry.sty} for margin settings;
%\item {fleqn.clo} for left aligned equations;
%\item {graphicx.sty} for graphics inclusion;
%\item {hyperref.sty} optional packages if hyperlinking is
%  required in the document;
%\end{enumerate}  
Currently, rotating machinery is widely used in aviation, aerospace, shipbuilding, automobile and other industrial fields, acting as the power source and support of many industrial systems \cite{zhao2021}. The rolling bearing is a key vulnerable part of rotating machinery, which is prone to failure under a harsh working environment and alternating load \cite{yan2018}. Therefore, the research of rolling bearing fault diagnosis is of great importance to ensure the safety and reliability of facilities \cite{shao2015}.

The ultimate goal of fault diagnosis is to recognize the status of the target component in the machine and thus determine whether the machine needs maintenance. Generally, existing fault diagnosis methods consist of two categories: model-based \cite{frank2000} approaches and data-driven \cite{gao2015a} methods. Model-based methods often require a lot of prior knowledge, making difficult to accurately establish the diagnosis model of composite components under complex conditions. Data-driven methods \cite{gao2015a} aim to convert the data provided by sensors into a parametric or non-parametric correlation model. Data-driven methods can effectively and rapidly process machinery signals, providing accurate diagnosis results with low requirement for prior expertise. Therefore, they are becoming more and more attractive with the development of various intelligent approach. The most common intelligent fault diagnosis method is currently developed on machine learning (ML) approaches such as $k$-nearest neighbor ($k$-NN) \cite{dong2016}, support vector machine (SVM) \cite{yan2018}, self-organized map (SOM) network \cite{xiao2019}, etc. Most intelligent fault diagnosis methods for rolling bearings are built on the processing and analysis of vibration signals \cite{zhao2021}, which employ a discrimination model with the input of man-made features extracted from acquired raw signals. Note that these man-made features include: 1) time domain statistic moment \cite{durkhure} such as root-mean-square (RMS), kurtosis, skewness, etc., 2) frequency spectrum \cite{seryasat2010} processed by fast Fourier transform (FFT), power spectrum estimation, etc., 3) time--frequency domain energy \cite{yu2006} obtained with empirical mode decomposition (EMD), variational mode decomposition (VMD), etc., and even 4) fusion features \cite{wang2014} extracted with principal component analysis (PCA), etc.

Recently, the development of deep learning (DL) allowed us to automatically learn representations from a large amount of data, thus avoiding the need for manual design features \cite{lecun2015}. Shao \emph{et al}. \cite{shao2015} proposed a rolling bearing fault diagnosis method based on a deep belief network (DBN). Mao \emph{et al}. \cite{mao2017} combined auto-encoder (AE) and extreme learning machine (ELM) to diagnose fault mode of rolling bearings, which utilizes FFT spectrum of vibration signals as input. Based on DL architecture, the convolutional neural network (CNN) which is specifically designed for variable and complex signals, has shown great merits in feature extraction. Xu \emph{et al}. \cite{xu2020} introduced VMD and deep convolutional neural networks (DCNN) to perform fault classification of the rolling bearing in wind turbines. Jia \emph{et al}. \cite{jia2018} employed a deep normalized convolutional neural network (NCNN) for imbalanced fault classification of machinery and analyzed its mechanism via visualization. Besides, recurrent neural network (RNN), another significant DL model, shows an advantage in learning internal features from the input of sequences, so it is also widely used in the field of diagnosis in dependence of time-series vibration signals. Liu \emph{et al}. \cite{liu2018} offered a fault diagnosis framework of rolling bearings with recurrent neural network (RNN) and auto-encoder (AE). To solve the problem of convergence difficulty and gradient extinction of common RNNs, Chen \emph{et al}. \cite{chen2017} introduced the long short-term memory network (LSTM), a variant of RNN, to the prediction of mechanical state. Zhao \emph{et al}. \cite{zhao2018} came up with an end-to-end fault diagnosis method based on LSTM neural network, which can directly classify the raw process data without specific feature extraction and classifier design. In addition to the above listed, a large number of fault diagnosis methods based on deep learning are constantly proposed \cite{shao2017,zhang2020}. CNN and RNN are the two most common architectures in the field of DL and DL-based fault diagnosis. However, these two types of neural networks have their own defects. For example, RNNs and their variants are not suitable for parallel computation, which is very inefficient for training on large-scale datasets. In addition, RNN variants still cannot completely avoid long-range dependence problem, that is, the difficulty in establishing an effective connection between distant sequences \cite{kolen2010}. As for CNN, another pillar of DL, it also suffers from some shortcomings such as the lack of capturing the relationships between targets, the equal treatment of all pixel points and lack of pertinence \cite{wu2020}. Besides, the local receptive field of convolutional kernel results in the need for numerous convolutional layers to be stacked in order to obtain global information \cite{gulati2020}.

Nowadays, attention mechanism, which can associate different positions of a sequence to compute its unique representation, has now been successfully applied in a variety of tasks including natural language processing (NLP), computer vision (CV), and even fault diagnose \cite{li2019a}. Long \emph{et al}. \cite{long2021} introduced a motor fault diagnosis method using attention mechanism and improved AdaBoost. Li \emph{et al}. \cite{li2019a} leveraged the attention mechanism to improve the data-driven diagnosis approach and visualized its effect on learned knowledge. The attention mechanism assists the deep network to focus more on informative data segments and ignore those that contribute less to the final output. However, all these attempts only embed attention mechanism as an auxiliary module into backbone models such as CNN or RNN, which cannot completely avoid the defects of these classical models.

To take full advantage of the attention mechanism, Vaswani \emph{et al}. \cite{vaswani2017} proposed a new architecture based only on attention mechanism -- Transformer, which abandons all the recurrent and convolutional structures. We call this version vanilla Transformer. The proposal of Transformer has set off a revolution in the field of NLP. So far, there have been numerous varieties of Transformer, and a large number of successful practices have been put into machine translation, sentence generation, etc \cite{shazeer2017a}. Computer vision and other fields are also attempting to introduce and improve Transformer to meet the new challenges \cite{dosovitskiy2020}. Considering that fault diagnosis often needs to process the signal sequence and extract its internal correlation, Transformer-like models should have yielded unusually brilliant results in this field. Besides, even when dealing with two-dimensional input such as time--frequency representation, Transformer is good at grasping its inherent temporal correlation. Unfortunately, having said that, Transformer has not been used in fault diagnosis and related fields. 

To better model the temporal information in bearing signals, construct long-distance dependence and extract more effective hidden representation from its time--frequency representation (TFR), a new model named time--frequency Transformer (TFT) is proposed in this paper. Constrained by the difficulty in extracting useful features directly from the raw vibration signal \cite{li2019a}, we apply synchrosqueezed wavelet transform (SWT) \cite{daubechies2011} to obtain time--frequency representations (TFRs). SWT has been applied to many prognostic and health management (PHM) studies of bearings due to their good performance in non-stationary vibration signal processing \cite{wen2016a,li2016a,yi2018,liu2020}. Then, TFT is proposed to provide a discriminate model between TFRs and bearing fault modes. We designed a novel tokenizer focused on TFRs, and an encoder composed of Transformer blocks to establish hidden representations. Thus, we proposed an end-to-end approach for fault diagnosis. Through the case studies on bearing experimental datasets, we constructed the optimal Transformer structure and verified the performance of the proposed diagnosis method. Through comparison with the benchmark models and other state-of-the-art methods, superiority of the proposed method is proved. The main contributions of this paper can be listed as follows.

\begin{enumerate}[1)]
	\item We proposed a novel time--frequency Transformer, which can avoid some drawbacks of classical models such as RNN and CNN, to extract effective information from time--frequency representation with only attention mechanism.
	\item We proposed an end-to-end fault diagnosis framework based on time--frequency Transformer and synchrosqueezed wavelet transform, which proves to be effective and superior on bearing experimental datasets. 
\end{enumerate}

%The rest of this paper is organized as follows. Section II introduces the vanilla Transformer and its existing variants. The proposed time--frequency Transformer and the fault diagnosis framework based on it are detailed in Sections III and IV respectively. In Section V, case studies on two bearing experimental datasets are described to validate the effectiveness of the proposed model and the fault diagnosis method. Section VI summarizes the article.

\section{Preliminaries}
\label{sec:preliminaries}
This section will briefly introduce the vanilla Transformer proposed by Vaswani \emph{et al}. \cite{vaswani2017} in 2017. Variants of Transformer applicated in the fields of natural language processing (NLP) and computer vision (CV) will also be reviewed.

\subsection{Transformer}
Recurrent models have shown a good capability to process sequence input in the form of $\left[x_1,x_2,\dots,x_t\right]$. Along the direction of the input sequence, they generate a sequence of hidden states $h_t$, as a function of the previous hidden state $h_{t-1}$ and the input token $x_t$ for position $t$. At each step the model is auto-regressive, consuming the previously generated symbols as additional input when generating the next. This inherently sequential nature precludes parallelization within training examples, which becomes critical at longer sequence lengths. Vaswani \emph{et al}. \cite{vaswani2017} proposed Transformer, a new architecture relying entirely on attention mechanism to draw global dependencies between input and output. Transformer completely abandons the traditional recurrent structure to realize the parallel calculation of sequence input. In addition, convolution operation which is difficult to globally model the relationship between local features is eliminated from in Transformer \cite{kolen2010}.

\begin{figure}[htb]
	\centerline{\includegraphics[width=\columnwidth]{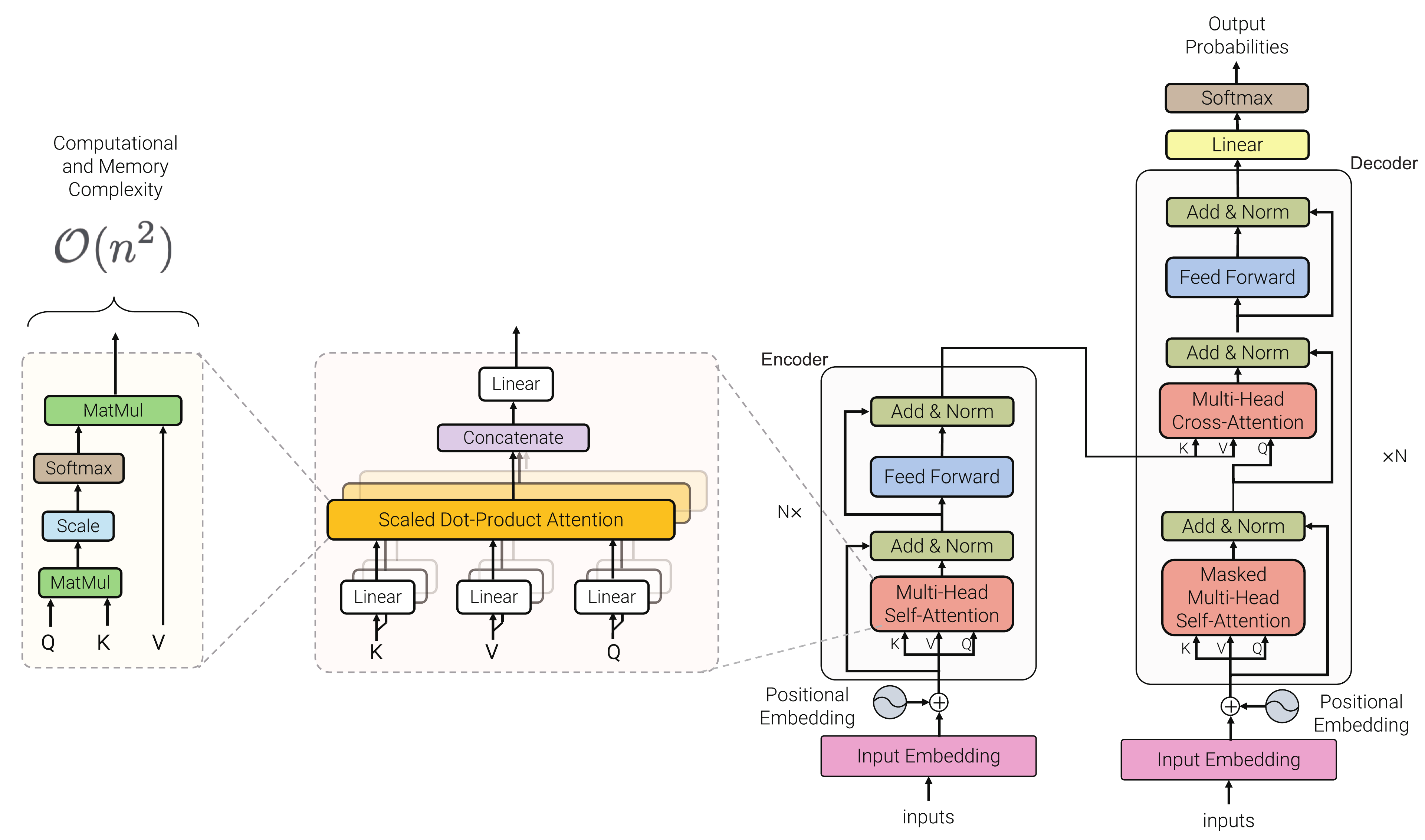}}
	\caption{Architecture of the vanilla Transformer \cite{vaswani2017}.}
	\label{fig:transformer}
\end{figure}

Transformer is a multilayer structure by stacking Transformer blocks, whose vanilla form is shown in Fig. \ref{fig:transformer}. The Transformer block is characterized by a multi-head self--attention mechanism, a position-wise feed-forward network, layer normalization \cite{ba2016} module and residual connector \cite{he2016}. The input to the Transformer is often a tensor of shape $\mathbb{R}^b \times \mathbb{R}^n$, where $b$ is the batch size, $n$ is the sequence length (note that the difference from the dimension of sequence). The input first passes through an embedding layer, which converts each one-hot token into an embedding of $d$ dimensions to obtain a new tensor, i.e., $\mathbb{R}^b \times \mathbb{R}^n \times \mathbb{R}^d$. Then, the new tensor is added to a sinusoidal position encoding, and passes through a multi-head self--attention module. The input and output of the multi-head self--attention are connected by a layer normalization layer and a residual connector. The combined output is then sent to a two-layer position-wise feed-forward network, which similarly connects the input and output through a residual connector and a layer normalization layer. Such sublayer residual connectors with layer normalization have the following form
\begin{equation}
	X_{out}={\rm LayerNorm}(F_{A/FF}(X_{in})+X_{in})
\end{equation}
where $F_{A/FF}$ denotes multi-head self--attention or position-wise feed-forward layers.

\subsubsection{Multi-head self--attention}
The multi-head self--attention mechanism is a key defining characteristic of Transformer models, whose mechanism behind can be viewed as learning an alignment, that is, each token in the sequence attempts to gather information from others \cite{bahdanau2015}. Generally, through linear transformations on a group of input embeddings $X$ with dimension $d_{\rm model}$, we get queries $Q_s=XW_s^q$ and keys $K_s=XW_s^k$ with dimension $d_k$, and values $V_s=XW_s^v$ with dimension $d_v$.  A single-head scaled dot-product attention calculates the dot products of all queries and keys, divides each by scaling factor $\sqrt{d_k}$, and apply a softmax function to obtain the weights on the values. 
\begin{equation}
	A_s\left(Q_s,K_s,V_s\right)={\rm softmax}\left(\frac{Q_sK_s^{\top}}{\sqrt{d_k}}\right)V_s
	\label{eq:mha}
\end{equation}

However, instead of mapping with only one version of linear transformations, it is more beneficial to project the input tokens to different queries, keys and values $h$ times through different learned linear transformations. Thus, the so-called multi-head self--attention is introduced. Through the parallel self--attention calculation on queries, keys, and values of each projected version, $h$ different outputs ${\rm head}_i$ are obtained. These ${\rm head}_i$ are concatenated and once again projected, resulting in the multi-head self--attention
\begin{equation}
	\begin{split}
		& A_h\left(X\right)={\rm concat}\left({\rm head}_1,\dots,{\rm head}_h\right)W^O \\
		& {\rm where\ \rm head}_i=A_s\left(XW_i^q,XW_i^k,XW_i^v\right)
	\end{split}
\end{equation}
where $W_i^q \in \mathbb{R}^{d_{\rm model} \times d_k}$, $W_i^k \in \mathbb{R}^{d_{\rm model} \times d_k}$ and $W_i^v \in \mathbb{R}^{d_{\rm model} \times d_v}$ denote $i$-th version of linear projection on embeddings $X$ to obtain different queries, keys and values, respectively. $W^O \in \mathbb{R}^{h \cdot d_v \times d_{\rm model}}$ denotes the linear projection on concatenated multi-head. Note that $d_k=d_v=d_{\rm model}/h$ in Transformer.

\subsubsection{Position-wise feed-forward layers}
The output of multi-head self--attention module is then passed through a two-layer feed-forward network, whose hidden layer is activated by ReLU. This feed-forward layer operates on each position independently hence the term position-wise.
\begin{equation}
	FF\left(X_A\right)={\rm ReLU}\left(0,X_AW_1+b_1\right)W_2+b_2
\end{equation}
where $W_1 \in \mathbb{R}^{d_{\rm model} \times d_{ff}}$, $W_2 \in \mathbb{R}^{d_{ff} \times d_{\rm model}}$, $b_1 \in \mathbb{R}^{d_{ff}}$, $b_2 \in \mathbb{R}^{d_{\rm model}}$  denotes the weights and bias of two layers, respectively.

\subsubsection{Transformer block}
Transformer composes of several stacked Transformer blocks. A Transformer block usually contains a multi-head self--attention module and a position-wise feed-forward module, both of which use a residual connector and layer normalization to get the combined output of the module.
\begin{equation}
	\begin{split}
		& X_A={\rm LayerNorm}\left(A_h\left(X\right)+X\right) \\
		& X_{FF}={\rm LayerNorm}\left(FF\left(X_A\right)+X_A\right)
	\end{split}
\end{equation}
where $X_A$ and $X_{FF}$ are the output of the multi-head self--attention module and the position-wise feed-forward module, respectively. Note that the stacked multiple Transformer blocks take the same structure, while do not share the same parameters.

\subsubsection{Transformer Mode}
It is important to note the differences in the mode of usage of the Transformer block. Transformers generally can be divided into three categories, named: 1) encoder-only (e.g., for classification), 2) decoder-only (e.g., for language modeling), and 3) encoder-decoder (e.g., for machine translation). The vanilla Transformer proposed by Vaswani uses an encoder-decoder structure (as shown in Fig. \ref{fig:transformer}) for machine translation, which is a Seq2Seq problem. The decoder of vanilla Transformer uses Transformer blocks that are different from the aforementioned those for the encoder. The proposed method in this paper, however, adopts an encoder-only structure, so details of decoders are not introduced.

\subsection{Transformer variants}
Recurrent Neural Networks, especially LSTM \cite{hochreiter1997} and GRU \cite{chung2015}, have been widely used for sequence modeling and inference problems such as machine translation and language modeling before Transformer was proposed \cite{shazeer2017a}. The proposal of Transformer brings brand new solutions to the NLP community. Devlin \emph{et al.} \cite{devlin2019} proposed the bidirectional Transformers (BERT), which is one of the most powerful models in the NLP field. Besides, models such as XLNet \cite{yang2020a} and GPT \cite{brown2020} further expand the application of Transformer. Tay \emph{et al.} \cite{tay2020} surveyed a dizzying number of efficient Transformer variants, providing an organized and comprehensive overview of existing work and models across multiple application fields.

Transformer is used for sequence modeling to solve several problems of traditional recurrent structures: 1) Difficulty in training parallelization. 2) Difficulty in modeling long-range dependencies. In many sequence transduction tasks, learning long-range dependence is a key challenge. The recurrent models cannot establish a direct connection between non-adjacent tokens, which greatly limits the overall understanding of the input sequence. 3) The problem of gradient explosion and gradient vanishes. The proposal of LSTM and GRU alleviated this problem but did not fundamentally eliminate the weight accumulation caused by multiple recurrences.

The great success of Transformer in NLP has also attracted the attention of researchers in the field of CV. For many years, convolutional neural network (CNN) is the fundamental pillar in CV. However, this convolution operating on the pixel matrix also has some defects, e.g., it is difficult to capture the relationship between targets, and treats all pixels equally without pertinence \cite{wu2020}. Many attempts \cite{wang2017a,hu2020,zhang2019a} introduce attention mechanisms into visual tasks. The trend of using Transformer as neural network module is more and more obvious. Dosovitskiy \emph{et al.} \cite{dosovitskiy2020} proposed a vision Transformer (ViT) for supervised image classification. It divides the image into several patches, and patches are treated the same way as tokens (words) in an NLP application. Wu \emph{et al.} \cite{wu2020} input convolution image features into Transformer as tokens, and used them for image classification and object detection. Guo \emph{et al.} \cite{guo2020} put forward a point cloud Transformer (PCT) for point cloud learning to solve 3D computer vision problems.

RNN and CNN, as the two most important network structures in DL, have been widely used in the field of bearing PHM. The vibration signals of bearings contain abundant temporal correlation, which can be well modeled by RNN to give effective fault diagnosis or residual useful life (RUL) prediction. CNN can extract useful fault or degradation information from the feature matrix composed of time, frequency and time--frequency domain features. A large number of model implementations based on CNN are also derived as reviewed in Section \ref{sec:introduction}. Sometimes, scholars also build a hybrid model combining these two structures. A large number of attempts broaden the application of DL in the field of bearing PHM. However, as mentioned above, these two models are now dwarfed for their inherent defects by the emergence of Transformer. To the best of our knowledge, no Transformer variant has been proposed to model the vibration signal and its TFRs, which can be useful for bearing PHM.

\section{Time--frequency Transformer}
\begin{figure}[htb]
	\centerline{\includegraphics[width=\columnwidth]{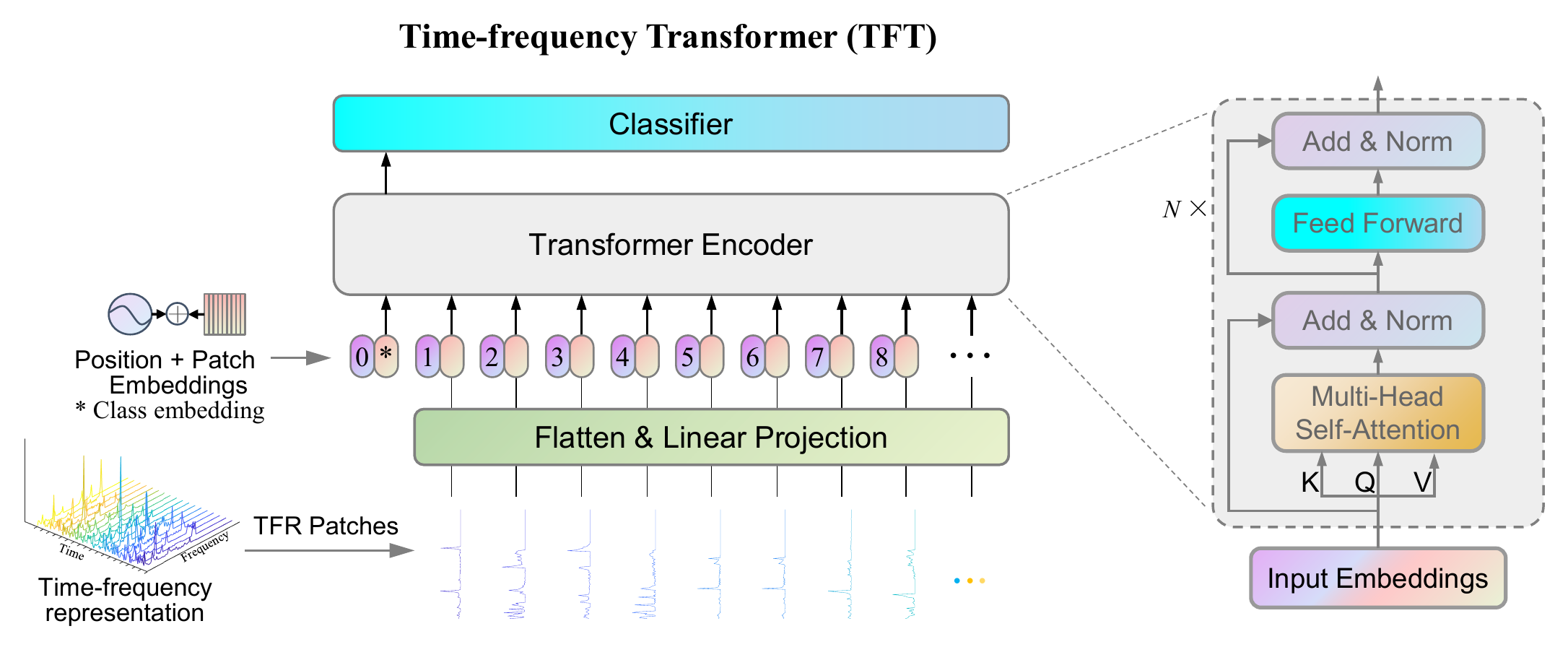}}
	\caption{The proposed time--frequency Transformer.}
	\label{fig:tft}
\end{figure}

This section will introduce the proposed time--frequency Transformer (TFT) in detail, whose overall diagram is shown in Fig. \ref{fig:tft}. The network architecture is mainly composed of a tokenizer, an encoder and a classifier.

\subsection{Tokenizer}
The vanilla Transformer accepts a 1D token sequence and obtains token embeddings with a tokenizer by dictionary query. However, the input token here to be processed is time--frequency representation (TFR), rather than the words that can be queried in a dictionary. Thus, a specific tokenizer needs to be designed. To process 2D TFR data, we will design a new tokenizer module, which mainly includes flattening, segmentation, linear mapping and adding position encodings.

\subsubsection{Token embedding}
To obtain the tokens sequence, TFR is segmented into several patches along the time direction. Specifically, given a TFR ${\rm x} \in \mathbb{R}^{N_t \times N_f \times C}$, where $N_t$ and $N_f$ are the length in time and frequency direction, respectively. $C$ is the number of channels, which generally refers to the stacked layers of multi-sensor signals. We first reshape $\rm x$ into sequence ${\rm x}^{\prime} \in \mathbb{R}^{N_t \times (N_f \cdot C)}$ to flatten multiple channels, and cut it along the time direction to get a patch sequence $\left[{\rm x}_p^1,{\rm x}_p^2,\dots,{\rm x}_p^{N_t}\right]$ with length $N_t$, where ${\rm x}_p^i \in \mathbb{R}^{N_f \cdot C}$, $i=1,\dots,N_t$ . Then, to obtain the sequence of token embeddings with dimension $d_{\rm model}$, a learnable linear transformation is used to obtain the projected patches sequence ${\rm x}_t$. This process is expressed as:
\begin{equation}
	\begin{split}
		& {\rm x} \stackrel{\rm reshape}{\longrightarrow} {\rm x}_p \\
		& {\rm x}_t={\rm x}_p W_t
	\end{split}
\end{equation}
where $W_t \in \mathbb{R}^{N_f \times d_{\rm model}}$ denotes the learnable linear mapping. Such processing is based on the view that TFR is formed by splicing the instantaneous spectrum of the signal over a period of time. This is different from gridding segmentation on images by ViT \cite{dosovitskiy2020}, because we believe that grid cutting TFR will not retain instantaneous spectrum estimation at a certain time in each patch. The segmented TFR patches can be regarded as a sequence of instantaneous spectrum in a period of time, and processing such temporal sequence is the strength of Transformer-based structure.

\subsubsection{Class token}
Similar to the class token in BERT \cite{devlin2019}, a randomly initialized trainable embedding ${\rm x}_t^0={\rm x}_{\rm class} \in \mathbb{R}^{d_{\rm model}}$ is added to the beginning of the embedded token sequence. Thus, an embedding sequence ${\rm x}_t=\left[{\rm x}_{\rm class};{\rm x}_p^1 W_t,{\rm x}_p^2 W_t,\dots,{\rm x}_p^{N_t} W_t\right]$ with length $N_t+1$ is obtained. Note that output ${\rm z}_N^0$ of the class token after processed by the subsequent Transformer encoder will serve as the hidden representation of TFR.

\subsubsection{Position encoding}
Since the Transformer contains no recurrence or convolutional operations, to make full use of the sequence order, we should inject some information about the relative or absolute position of the tokens into the embedding sequence. Vanilla Transformer employs a kind of sinusoid position encoding based on corpus dictionary and token location \cite{vaswani2017}, which is not suitable for the problem we are trying to solve. We propose to use a learnable position encoding $E_{\rm pos} \in \mathbb{R}^{(N_t+1) \times d_{\rm model}}$ to extract position information more flexibly through the learning process. The position encodings and token embeddings are added to get the input embeddings
\begin{equation}
	{\rm z}_0=\left[{\rm x}_{\rm class};{\rm x}_p^1 W_t,{\rm x}_p^2 W_t,\dots,{\rm x}_p^{N_t} W_t\right]+E_{\rm pos}
\end{equation}
Two types of learnable position encodings, 1D and 2D, are considered. 1) The 1D position encoding $E_{\rm 1d \ pos} \in \mathbb{R}^{(N_t+1) \times d_{\rm model}}$ is broadcasted from a vector $e_{\rm 1d} \in \mathbb{R}^{N_t+1}$, i.e., $E_{\rm 1d \ pos}={\rm broadcast}(e_{\rm 1d})$. 1D position encoding can only encode the relative and absolute position information among $N_t+1$ tokens since the vector elements in the same token share the same encoding. 2) 2D position encoding $E_{\rm 2d \ pos} \in \mathbb{R}^{(N_t+1) \times d_{\rm model}}$ is a learning matrix with dimensions $(N_t+1) \times d_{\rm model}$, so it can simultaneously encode the location information between and within the tokens. Subsequent case studies will analyze the performance of these two encodings.

\subsection{Encoder}
In TFT, the encoder can be regarded as a feature extraction structure, which undertakes the task of mining category related information from the input embeddings sequence. The Transformer encoder, which is composed of $N$ Transformer blocks, takes the embedded sequence ${\rm z}_0$ as the input. Our TFT model basically employs the vanilla Transformer block structure, that is, multi-head self--attention module and position-wise feed-forward layers with residual connector and layer normalization, which has been described in Section \ref{sec:preliminaries}. Transformer block and multi-head self--attention mechanism inside are the key defining characteristics of Transformer-like models. Besides, we also made the following improvements.
\subsubsection{GeLU activation}
To improve the convergence of the network, we adopt Gaussian error linear units (GeLU) activation \cite{hendrycks2016} in feed-forward layers instead of ReLU activation used in vanilla Transformer. GeLU is defined as the product of input $x$ and mask $m \sim {\rm Bernouli(\Phi(x))}$, where $\Phi (x)=P(X \leq x)$, $X \sim \mathcal{N}(0,1)$ is the cumulative distribution function of the standard normal distribution. This distribution is chosen since neuron inputs tend to follow a normal distribution, especially with layer normalization. In this setting, inputs have a higher probability of being “dropped” as $x$ decreases, so the transformation applied to $x$ is stochastic yet depends upon the input. 

\begin{equation}
	{\rm GeLU}(x)=xP\left(X \leq x\right)=x \Phi \left(x\right)=x \cdot \frac{1}{2} \left[1+ erf\left(x/\sqrt{2}\right) \right]
\end{equation}

\begin{figure}[htb]
	\centerline{\includegraphics[width=0.35\columnwidth]{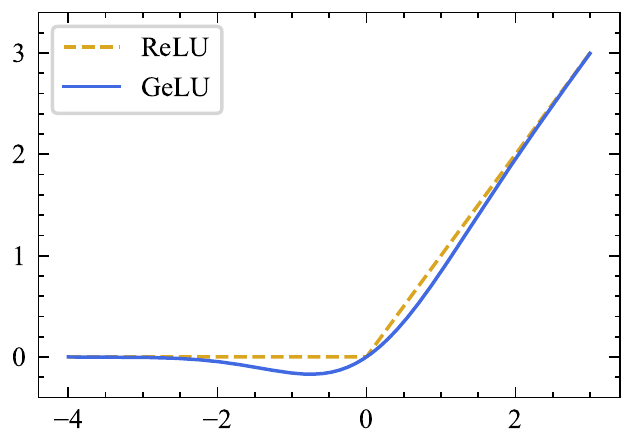}}
	\caption{The ReLU and GeLU ($\mu =0, \sigma =1$).}
	\label{fig:gelu}
\end{figure}

We can estimate GeLU as
\begin{equation}
	0.5x\left(1+{\rm tanh} \left[\sqrt{2/\pi}\left(x+0.044715x^3\right)\right]\right)
\end{equation}
if greater feedforward speed is worth the cost of exactness. GeLU and ReLU are plotted in Fig. \ref{fig:gelu}. It can be seen that GeLU activation is continuously differentiable and has more obvious nonlinearity than the non-differentiable ReLU activation at $x=0$.

\subsubsection{Transformer blocks in TFT}
The embedding sequence ${\rm z}_0=\left[{\rm z}_0^0;{\rm z}_0^1;\dots;{\rm z}_0^{N_t} \right]$ obtained by tokenizer goes through multiple Transformer blocks to extract the connection among the tokens using self--attention mechanism. An encoder contains $N$ Transformer blocks can be represented as
\begin{equation}
	\begin{split}
		& {\rm z}_l^{\prime}={\rm LayerNorm}\left(A_h\left({\rm z}_{l-1} \right)+{\rm z}_{l-1}  \right), l=1,\dots,N \\
		& {\rm z}_l={\rm LayerNorm}\left(FF\left({\rm z}_{l-1}^{\prime} \right)+{\rm z}_{l-1}^{\prime}  \right), l=1,\dots,N
	\end{split}
\end{equation}
where $A_h(\cdot)$ denotes the multi-head self--attention function with $h$ heads in Eq. (\ref{eq:mha}). $FF(\cdot)$ denotes the position-wise feed-forward module with $d_{\rm model}$ dimensions input and output, and $d_{ff}$ dimensions hidden layer. That is, the multi-head self--attention modules and the feed-forward modules are alternately connected. The extensive use of residual connectors and layer normalization reduces the training difficulty of deep network, which is conducive to faster and more stable convergence.

In TFT, the class token used to output classification features is regarded as the next token to be predicted, that is, the output of token characterizing category is regarded as an auto-regressive prediction problem. This kind of setting is born out of the idea that obtaining the information needed by the words to be predicted from the front words when Transformer is first used in NLP tasks. When applied to classification problems, this solution also enables Transformer to better establish the relationship between the input sequence and the class token to be predicted. Thus, the hidden features ${\rm z}_N^0$ of the class token output through each layer of the encoder can contain sufficient information in the input sequence. In the last layer of the encoder, the hidden features ${\rm z}_N^0$ will be served as the output of the encoder for subsequent classification.

\subsection{Classifier}
The function of classifier is to map the hidden features with category information to one-hot encoding of class labels. The classifier consists of two layers of feed-forward multi-layer perceptron (MLP) and a softmax activation,
\begin{equation}
	{\rm CLA}\left({\rm z}_N^0 \right) ={\rm softmax}\left( {\rm GeLU}\left( {\rm z}_N^0 W_{c1}+b_{c1}\right) W_{c2}+b_{c2}\right) 
\end{equation}
where $W_{c1} \in \mathbb{R}^{d_{\rm model} \times d_{ff}}$, $W_{c2} \in \mathbb{R}^{d_{\rm model} \times N_{cla}}$, $b_{c1} \in \mathbb{R}^{d_{ff}}$, $b_{c2} \in \mathbb{R}^{N_{cla}}$ are the weights and biases of the two layers, respectively. $N_{cla}$ is the number of categories. To reduce the number of hyperparameters, the hidden layer uses the same hidden dimension $d_{ff}$ as in encoder.

Generally speaking, we can easily know the probability of each category according to the softmax output of the network. Thus, the category of input samples can be predicted.

\subsection{Training of TFT}
Training of TFT follows the general deep learning scheme that using stochastic gradient descent (SGD) and error back-propagation (BP) algorithm to minimize the empirical risk. Given a training set $\mathcal{D}=\left\lbrace {\rm x}_i,{\rm y}_i\right\rbrace _{i=1}^n$ contains $n$ samples, the network adopts the cross-entropy (CE) loss function which is suitable for the classification problem
\begin{equation}
	\mathcal{J}\left(\theta \right) =\frac{1}{n}\sum_{i=1}^n \mathcal{L}_{CE}\left({\rm y}_i,\hat{\rm y}_i \right) 
\end{equation}
where ${\rm y}_i$ and $\hat{\rm y}_i$ are the expected and estimated output of sample $x_i$. $\theta$ denotes the trainable parameters of TFT and $\mathcal{L}_{CE}\left(\cdot \right) $ denotes the cross-entropy loss function.
\begin{enumerate}[1)]
	\item \textbf{Optimizer}: We use Adam optimizer \cite{kingma2015} for TFT training. Adam wields an adaptive gradient strategy to accelerate training error convergence and allow the training trajectory to cross non-smooth regions of the loss landscape \cite{zhang2019b}.
	\item \textbf{Regularization}: We introduce dropout \cite{srivastava2014} into each sub-module of the network, which only plays a role in the training stage. By cutting off the connections of some neurons, dropout forces the network to learn more robust parameters to reduce overfitting. In addition, we use label smoothing \cite{szegedy2016}, mainly through soft one-hot to add noise, which reduces the weight of the real sample label category in the calculation of the loss function, to suppress the over-fitting effect.
\end{enumerate}

The detailed TFT training steps are shown in Algorithm \ref{al:tft}.

\begin{algorithm}[htb]
	\caption{Training of time--frequency Transformer.}%算法名字
	\label{al:tft}
	\LinesNumbered %要求显示行号
	\footnotesize
	\KwIn{Training set $\mathcal{X}=\left\lbrace {\rm x}_{i},{\rm y}_{i}\right\rbrace _{i=1}^{n_{S}}$, where ${\rm x}_{i}\in\mathbb{R}^{N_t \times N_f \times C} $, batch size $n_{b}$}%输入参数
	
	Initialize $\left\lbrace W^{(l)},b^{(l)}\right\rbrace $ of TFT\;
	\For{epoch=\rm 1,2,\dots,max\_epoch}{
		\For{step=\rm 1,2,\dots,max\_step}{
			//Tokenizer
			
			\For{\rm each ${\rm x}_i$ in $\left\{{\rm x}_{i}\right\}_{i=1}^{n_b}$}{
				
				Flatten and slice ${\rm x}_i$ into patches sequence, obtain ${\rm x}_{i,p}=\left[{\rm x}_{i,p}^1W_t;\ldots;{\rm x}_{i,p}^{N_t}W_t\right]$\;
				Linear projection, obtain $\left[{\rm x}_{i,t}^1;\ldots;{\rm x}_{i,t}^{N_t}\right]$\;
				Add class token, obtain ${\rm x}_{i,t}=\left[{\rm x}_{\rm class};{\rm x}_{i,p}^1W_t;\ldots;{\rm x}_{i,p}^{N_t}W_t\right]$\;
				Add position encoding, obtain ${\rm z}_{i,0}={\rm x}_{i,t}+E_{pos}$\;
				Stack batches, obtain input sequences ${\rm Z}_o$\;
			}
			// Encoder
			
			\For{{\rm block} l={\rm 1,\dots,}N}{
				
				${\rm Z}_l^{\prime}={\rm LayerNorm}\left(A_h\left({\rm Z}_{l-1}\right)+{\rm Z}_{l-1}\right)$\;
				${\rm Z}_l={\rm LayerNorm}\left(FF\left({\rm Z}_{l}^{\prime}\right)+{\rm Z}_{l}^{\prime}\right)$\;
			}
			// Classifier
			
			Obtain hidden representation of class token ${\rm Z}_N^0$\;
			$\hat{{\rm Y}}={\rm softmax}\left({\rm GeLU}\left({\rm Z}_N^0W_{c1}+b_{c1}\right)W_{c2}+b_{c2}\right)$\;
			
			// Calculate loss and gradient descent
			
			Cross entropy $\mathcal{L}_{CE}\left({\rm Y},\hat{\rm Y}\right)$\;
			Batch loss $\mathcal{J}\left({\rm Y},\hat{\rm Y};W^{(l)},b^{(l)}\right)$\;
			Calculate gradients $\frac{\partial \mathcal{J}}{\partial W^{(l)}}$,$\frac{\partial \mathcal{J}}{\partial b^{(l)}}$\;
			Apply gradients $W^{(l)}\leftarrow W^{(l)}-\eta \frac{\partial \mathcal{J}}{\partial W^{(l)}}$,$b^{(l)}\leftarrow b^{(l)}-\eta \frac{\partial \mathcal{J}}{\partial b^{(l)}}$\;	
		}
	}
	\KwOut{Weights and biases $\left\lbrace W^{(l)},b^{(l)}\right\rbrace $}%输出
\end{algorithm}

\section{Fault diagnosis framework based on TFT}
\begin{figure}[htb]
	\centerline{\includegraphics[width=0.65\columnwidth]{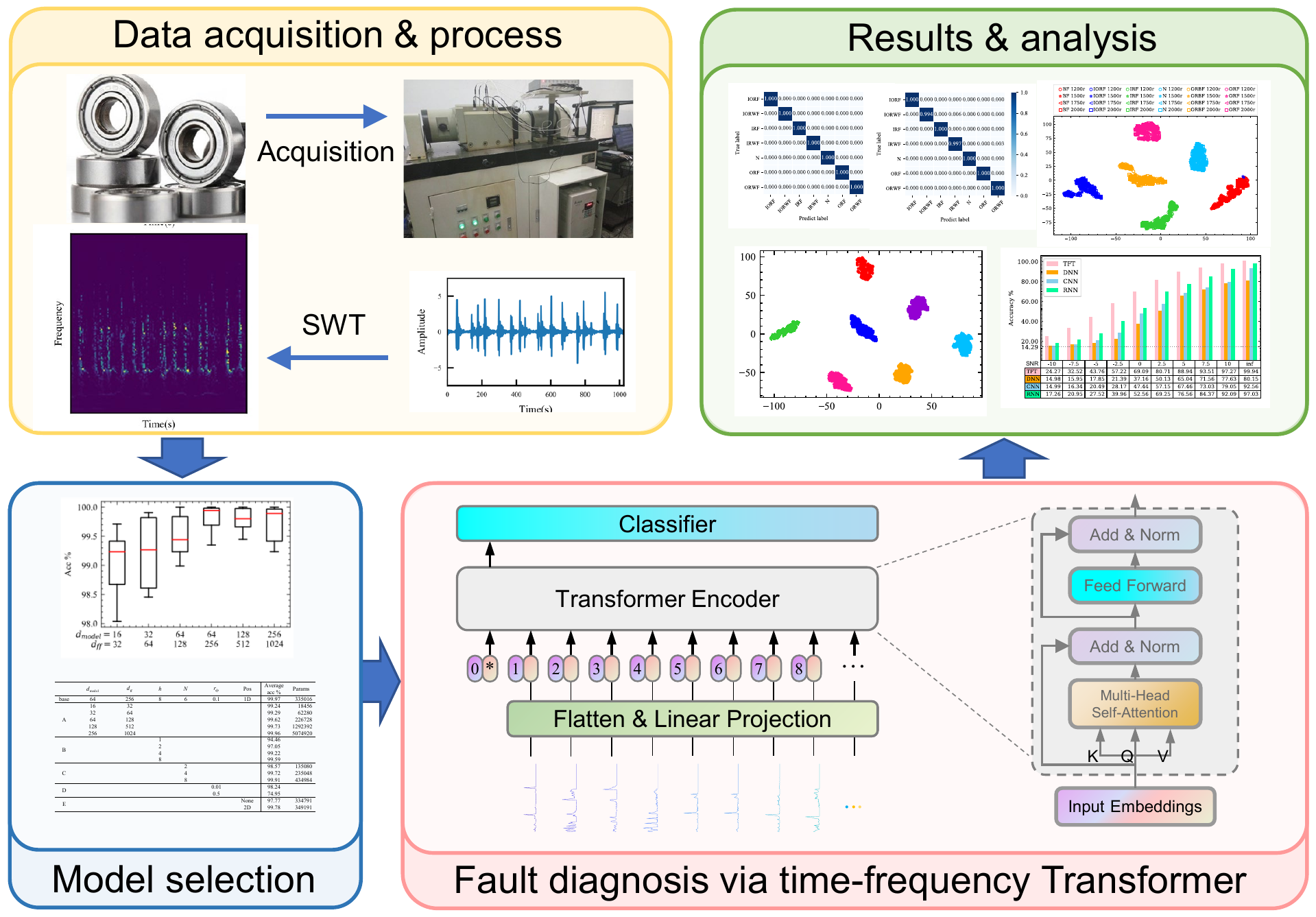}}
	\caption{The framework of the proposed method.}
	\label{fig:framework}
\end{figure}

To improve the generalization performance of intelligent fault diagnosis and speed up learning and inference, a new fault diagnosis method of rolling bearings based on TFT is proposed. Specifically, the process framework of the fault diagnosis method based on TFT is shown in Fig. \ref{fig:framework}, its specific diagnostic steps can be described as follows: 
\begin{enumerate}[1)]
	\item Collecting vibration signals from the rolling bearings. 
	\item Converting the collected vibration signals into TFRs through synchrosqueezed wavelet transforms (SWT), and label the samples for training. 
	\item The relevant hyperparameters and model structure of the TFT are determined. 
	\item The established model is fully trained and then applied to identify test samples. 
	\item Outputting the fault diagnosis results and evaluating the performance of the proposed method.
\end{enumerate}

\section{Case studies and analysis}
\begin{figure}[htb]
	\centerline{\includegraphics[width=0.4\columnwidth]{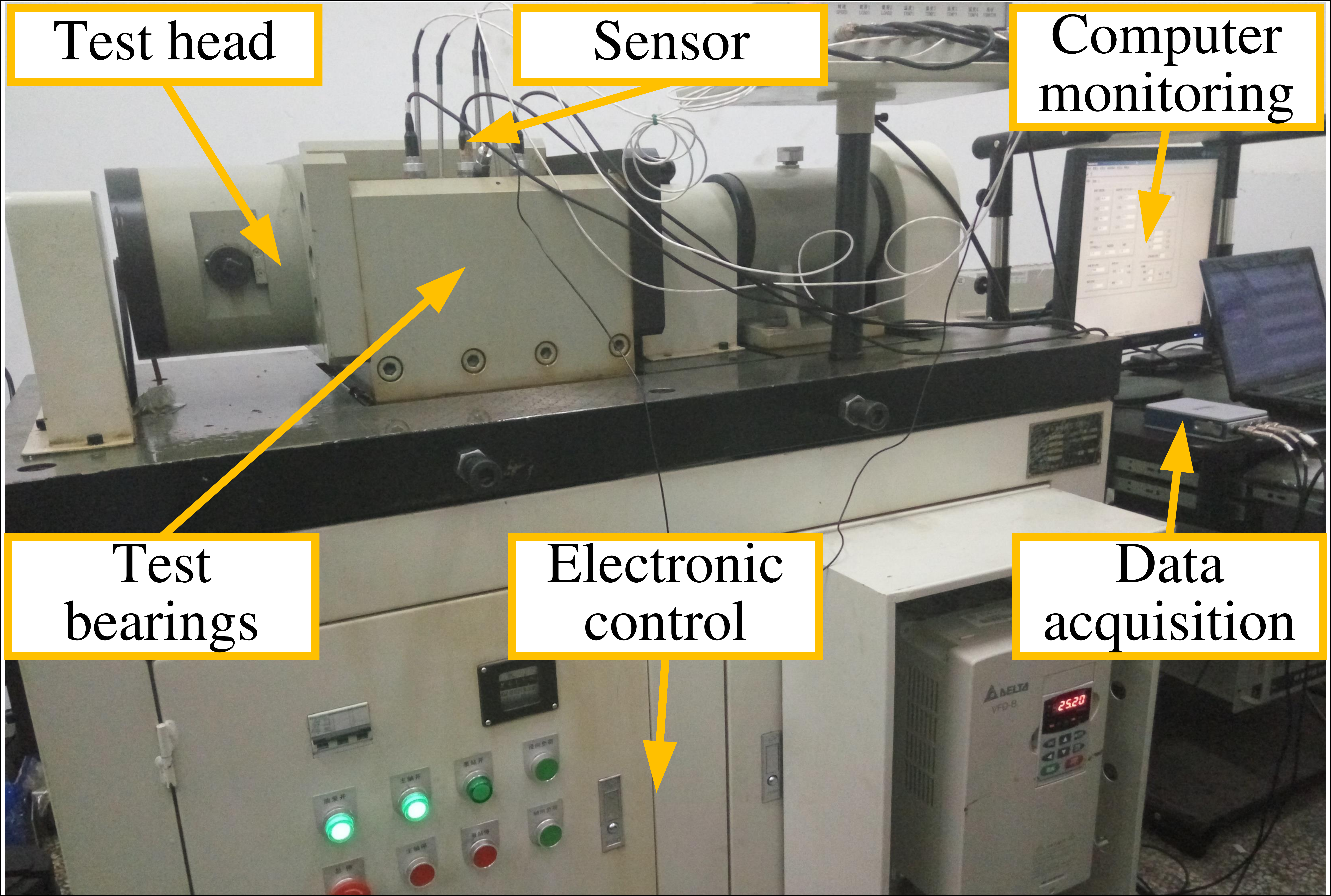}}
	\caption{ABLT-1A.}
	\label{fig:ablt}
\end{figure}

To verify and analyze the effectiveness of the proposed TFT and its application in bearing fault diagnosis, we implement two case studies in this section. The two bearing datasets used are collected from accelerated bearing life tester (ABLT-1A). The testing system is mainly composed of a test head, sensors, test bearings, electronic control system, computer monitoring system and data acquisition system. The specific real scene is shown in Fig. \ref{fig:ablt}. In our implementation, MATLAB is used for signal processing, while TensorFlow 2.1 framework is used for deep learning. All programs run on a computer with the following configuration: AMD Ryzen 2600, NVIDIA RTX 1060, 16GB RAM.

In the following case studies, we use three most popular deep learning models as benchmark models, as follows: 1) Multilayer perceptron (MLP) \cite{rumelhart1988}, as known as deep neural network (DNN). 2) Convolutional neural networks (CNN). To improve the convergence performance of deep convolutional networks, we employ ResNet18 \cite{he2016} with residual connection. 3) Recurrent neural network (RNN). To alleviate the long-range dependence problem of classical RNN, we use the improved GRU \cite{chung2015}. 

\subsection{Case 1: ABLT-1A Bearing Dataset 6308}
\subsubsection{Dataset description}
The rolling bearing HRB6308 is selected as the experimental bearing in this experiment. The faulty bearing is installed in the first channel of the sensor, and the other three normal bearings are installed in the remaining channels of the sensor. The vibration signal of the fault or normal rolling bearing is collected by the monoaxial vibration acceleration sensor with a sampling frequency of 12,800Hz. The experimental dataset is detailed as follows. Under zero-load conditions, seven types of fault conditions such as normal (6308N), inner ring fault (6308IRF), inner ring weak fault (6308IRWF), outer ring fault (6308ORF), outer ring weak fault (6308ORWF), inner and outer ring compound fault (6308IORF), and inner and outer ring compound weak fault (6308IORWF) are simulated. Accordingly, vibration data were collected for each failure type of the bearings, which were running at 1,050 rpm. For each failure mode interception, 2000 samples of the length 1024 were obtained, totaling $2000 \times 7 = 14 000$ samples. In addition, 60\% of all datasets are used as training dataset, 20\% as validation dataset for model selection and cross validation, and 20\% as test dataset for final test. In each training and test, the datasets are randomly divided to ensure the comprehensive evaluation of the model performance.

\subsubsection{Data preprocessing}
\begin{figure}[htb]
	\centering  
	\subfigtopskip=2pt %设置子图与上面正文或别的内容的距离
	\subfigbottomskip=0pt %设置第二行子图与第一行子图的距离，即下面的头与上面的脚的距离
	\subfigcapskip=0pt %设置子图与子标题之间的距离
	\subfigure[]{
		\label{fig:sig.a}
		\includegraphics[width=0.2\columnwidth]{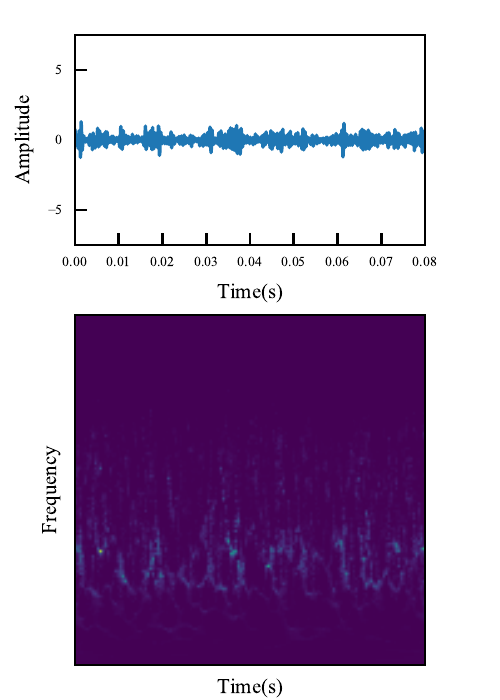}}
	\subfigure[]{
		\label{fig:sig.b}
		\includegraphics[width=0.2\columnwidth]{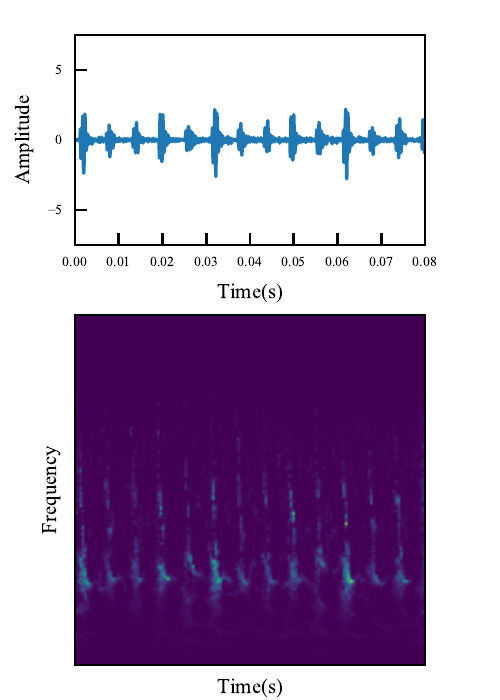}}
	\subfigure[]{
		\label{fig:sig.c}
		\includegraphics[width=0.2\columnwidth]{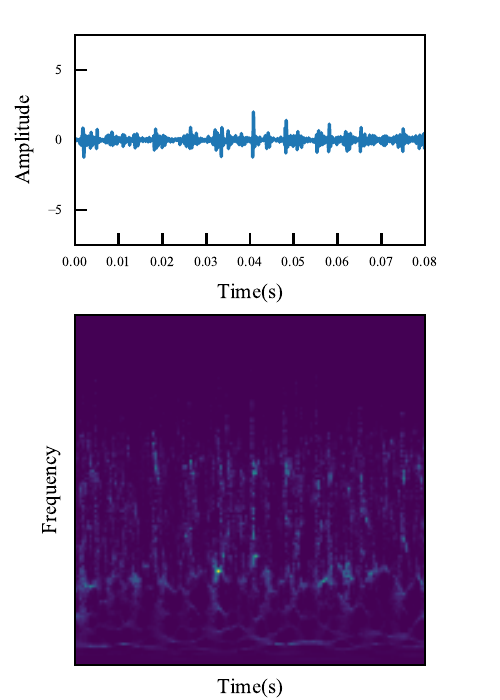}}
	\subfigure[]{
		\label{fig:sig.d}
		\includegraphics[width=0.2\columnwidth]{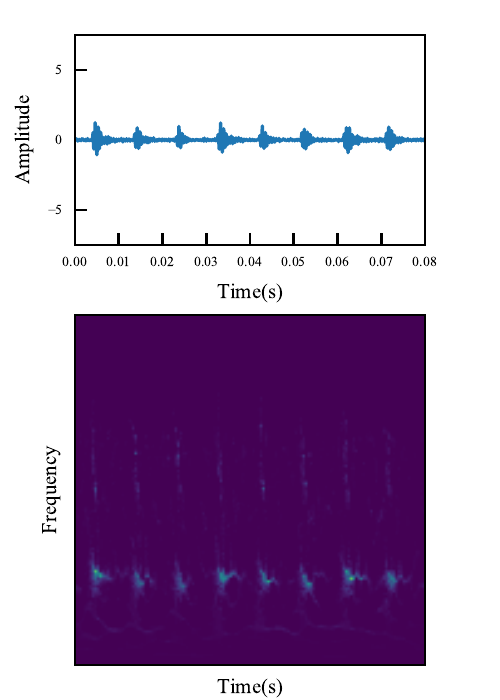}}
	\subfigure[]{
		\label{fig:sig.e}
		\includegraphics[width=0.2\columnwidth]{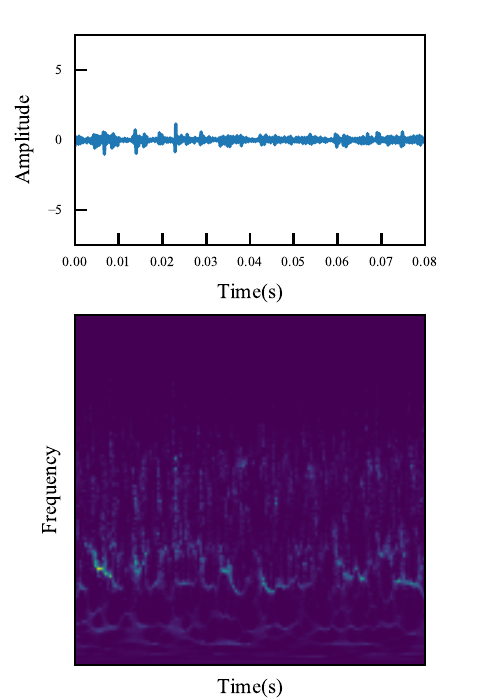}}
	\subfigure[]{
		\label{fig:sig.f}
		\includegraphics[width=0.2\columnwidth]{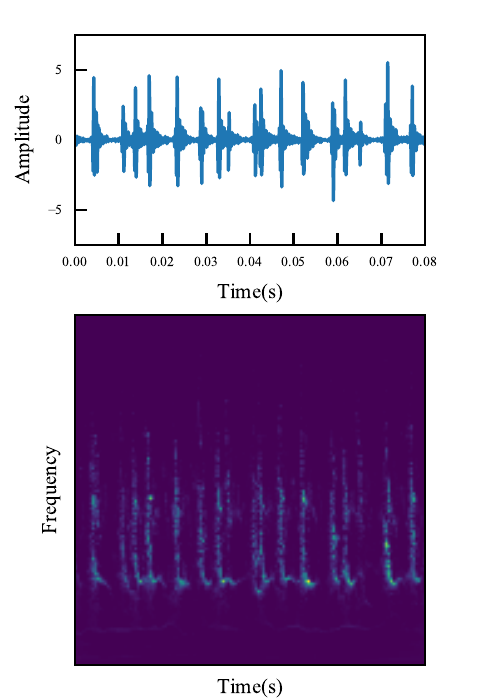}}
	\subfigure[]{
		\label{fig:sig.g}
		\includegraphics[width=0.2\columnwidth]{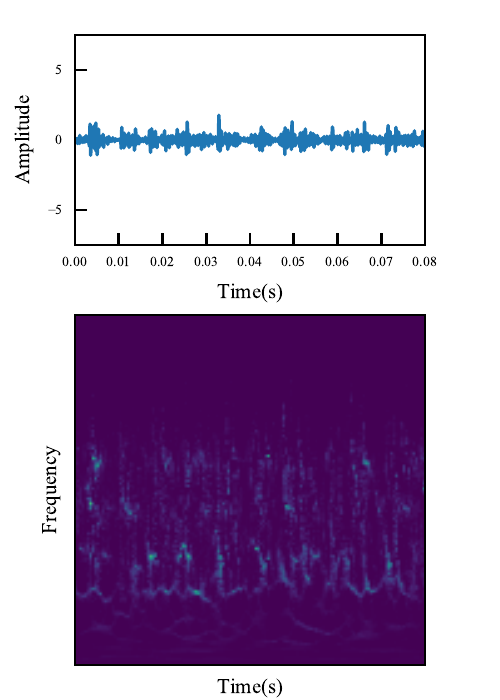}}
	\caption{Signals and SWT results of bearings (a) 6308N (b) 6308IRF (c) 6308IRWF (d) 6308ORF (e) 6308ORWF (f) 6308IORF (g) 6308IORWF.}
	\label{fig:sigs}
\end{figure}

While the raw vibration signal contains sufficient health condition information of bearings, it is not clear to be used for fault diagnosis directly \cite{li2019a}. Considering the non-stationarity of vibration signals, synchrosqueezed wavelet transforms (SWT) is used to process the raw data to obtain TFRs. The resolution of TFRs obtained by SWT is $1280 \times 2560$, which is too large to input a network. This will significantly increase scale of the network and computational expense. Similar to Ref. \cite{ding2021f}, the bicubic interpolation algorithm \cite{keys1981} is used to resize the resolution to $224 \times 224$, which is the common input size. Finally, input images with a shape of $224 \times 224 \times 1$ are obtained. Vibration signals and corresponding TFRs of bearings with different fault modes are drawn in Fig. \ref{fig:sigs}. These TFRs will be the input of the proposed fault diagnosis model.

\subsubsection{Model selection}
In this section, we will determine the structure and hyperparameter selection of the model based on the cross-validation on ABLT-1A Bearing Dataset 6308, and evaluate the influence of these hyperparameter selections on the model size and generalization performance. To increase the feasibility of the research work, all parameters and diagnostic results are cross-validated for multiple average verification and analysis. The hyperparameters we need to determine include: (A) embedding dimension $d_{\rm model}$ and hidden dimension $d_{ff}$, (B) number of attention heads $h$, (C) number of Transformer blocks $N$, (D) dropout rate $r_{dp}$, and (E) selection of position encoding. Each model was trained 10 times for cross-validation. Specifically, the average of the results of cross-validation is employed. The test results of each group are shown in Table \ref{tab:hyperparameters}, where base denotes the final selected model for subsequent research, average acc denotes mean testing accuracy, and params denotes the total number of trainable parameters in the model. The empty item in the params column indicates that this hyperparameter does not affect the total number of trainable parameters. It can be seen from the table that the selection of these parameters has a certain impact on the model scale and performance. In particular, different dimensions and encoder layers will greatly affect the scale of the model.

\begin{table}[htb]
	\caption{Model selection and influence of some hyperparameters.}
	\label{tab:hyperparameters}
	\footnotesize
	\begin{tabularx}{\columnwidth}{p{0.8cm}<{\centering}p{1.2cm}<{\centering}p{1.2cm}<{\centering}p{1.cm}<{\centering}p{1cm}<{\centering}p{1.2cm}<{\centering}p{2.2cm}<{\centering}|p{2cm}<{\centering}r}
		\hline
		& $d_{\rm model}$   & $d_{ff}$    & $h$ & $N$ & $r_{dp}$ & Position coding &  Average acc \% &  Params num \\ \hline
		base               & 64  & 256  & 8 & 6 & 0.1  & 1D   & 99.94 & 335,016   \\ \hline
		\multirow{5}{*}{A} & 16  & 32   &   &   &      &      & 99.14 & 18,456    \\
		& 32  & 64   &   &   &      &      & 99.21 & 62,280    \\
		& 64  & 128  &   &   &      &      & 99.52 & 226,728   \\
		& 128 & 512  &   &   &      &      & 99.73 & 1,292,392 \\
		& 256 & 1024 &   &   &      &      & 99.93 & 5,074,920 \\ \hline
		\multirow{4}{*}{B} &     &      & 1 &   &      &      & 94.46 &           \\
		&     &      & 2 &   &      &      & 97.05 &           \\
		&     &      & 4 &   &      &      & 99.22 &           \\
		&     &      & 16 &   &      &      & 99.59 &           \\ \hline
		\multirow{3}{*}{C} &     &      &   & 2 &      &      & 98.57 & 135,080   \\
		&     &      &   & 4 &      &      & 99.72 & 235,048   \\
		&     &      &   & 8 &      &      & 99.91 & 434,984   \\ \hline
		\multirow{2}{*}{D} &     &      &   &   & 0.01 &      & 98.24 &           \\
		&     &      &   &   & 0.5  &      & 74.95 &           \\ \hline
		\multirow{2}{*}{E} &     &      &   &   &      & None & 97.77 & 334,791   \\
		&     &      &   &   &      & 2D   & 99.78 & 349,191   \\ \hline
	\end{tabularx}
\end{table}

\begin{figure}[htb]
	\centering  
	\subfigcapskip=-8pt %设置子图与子标题之间的距离
	\subfigure[]{
		\includegraphics[width=0.35\columnwidth]{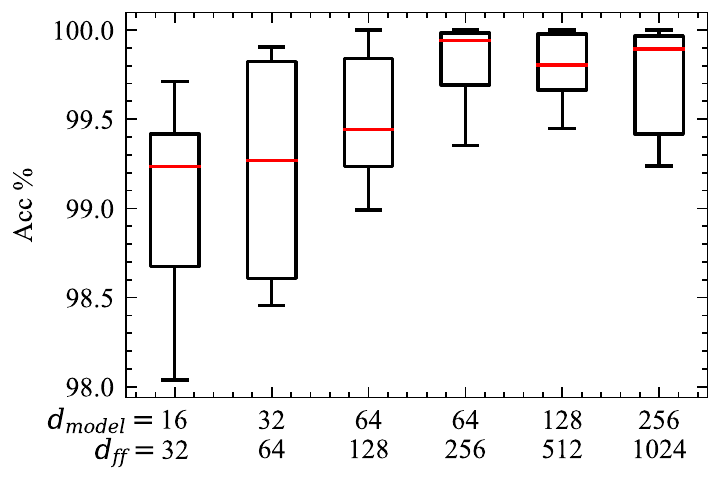}}
	\subfigure[]{
		\includegraphics[width=0.35\columnwidth]{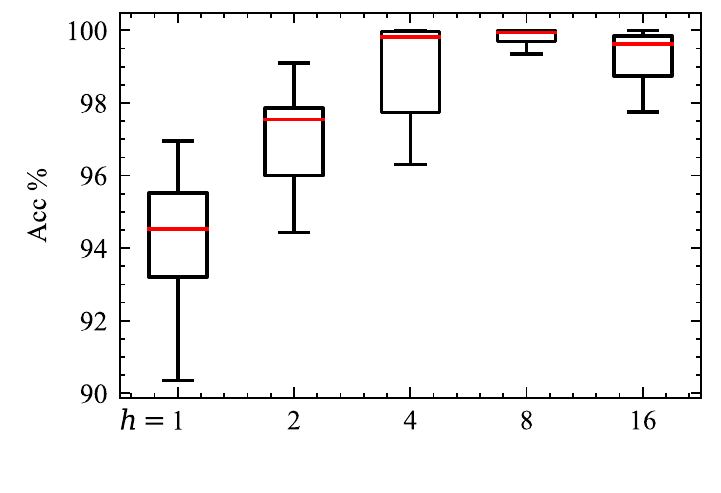}}
	\caption{The influence (a) embedding dimension and hidden dimension (b) the number of attention heads on the accuracy of time--frequency Transformer. }
	\label{fig:hyperparameter}
\end{figure}

In addition, for the embedding dimension $d_{\rm model}$, hidden layer dimension $d_{ff}$ and the number of attention heads $h$ that have a great impact on performance, the statistical box chart of test results is shown in Fig. \ref{fig:hyperparameter}. Too small embedding dimension and hidden layer dimension cannot provide the network with enough parameterization ability, so the network performance is poor. In contrast, too large dimension will lead to over-parameterization of the network. At this time, our samples will not be enough to train the over-parameterized network, which will lead to the degradation of the generalization performance. This is also called over-fitting. Besides, we notice that unreasonable selection of hyperparameters will not only reduce the average accuracy, but also make the network performance more unstable, that is, the error distribution is more dispersed. The number of attention heads also shows a similar pattern. Finally, the optimal network structure and hyperparameter selection of TFT are shown in Table \ref{tab:optimalHyperparameter}.

\begin{table}[htb]
	\centering
	\caption{Optimal structure and hyperparameters of the proposed time--frequency Transformer.}
	\label{tab:optimalHyperparameter}
	\footnotesize
	\begin{tabularx}{\columnwidth}{p{10cm}l}
		\hline
		& Value             \\\hline
		Input size								& {[}224, 224, 3{]} \\
		Batch size								& 32                \\
		Max epochs								& 100               \\
		Learning rate $lr$						& 5e-5              \\
		Optimizer								& Adam              \\
		Label smoothing rate $\epsilon _{ls}$	& 0.1               \\
		Num of encoder layers $N$				& 6                 \\
		Embedding dimension $d_{\rm model}$		& 64                \\
		Hidden dimension $d_{ff}$				& 256               \\
		Num of attention heads $h$				& 8                 \\
		Dropout rate$r_{dp}$					& 0.1               \\
		Position encoding						& 1D        		\\\hline       
	\end{tabularx}
\end{table}

For a fair comparison, the parameter settings of the three benchmark models are also standardized. The detailed model structure and hyperparameter settings are shown in Table \ref{tab:structure}.

\begin{table}[htb]
	\centering
	\caption{Structure and hyperparameter setting of three benchmark models.}
	\label{tab:structure}
	\footnotesize
	\begin{tabularx}{\columnwidth}{p{2cm}<{\centering}p{8cm}l}
		\hline
		Model & Structure (units and activation)                                                                                                                                            & Hyperparameter                                                                                                           \\\hline
		DNN   & \begin{tabular}[c]{@{}l@{}}Dense (512, activation='ReLU')\\ Dropout ( )\\ Dense (128, activation='ReLU')\\ Dropout ( )\\ Dense (num\_class, activation='ReLU')\end{tabular} & \begin{tabular}[c]{@{}l@{}}Dropout rate $r_{dp}$  \\ Max epochs = 100\\ Batch size = 32\\ Optimizer = Adam($lr=$1e-5)\end{tabular}  \\
		CNN   & \begin{tabular}[c]{@{}l@{}}ResNet18 ( )\\ GlobalAveragePooling2D ( )\\ Dense(num\_class)\end{tabular}                                                                       & \begin{tabular}[c]{@{}l@{}}Dropout rate $r_{dp}$ \\ Max epochs = 100\\ Batch size = 16\\ Optimizer = Adam ($lr=$5e-5)\end{tabular} \\
		GRU   & \begin{tabular}[c]{@{}l@{}}GRU (224, dropout) × 6\\ Dense (128, activation='ReLU')\\ Dense (num\_class, activation='ReLU')\end{tabular}                                     & \begin{tabular}[c]{@{}l@{}}Dropout rate $r_{dp}$ \\ Max epochs = 100\\ Batch size = 32\\ Optimizer = Adam ($lr=$5e-5)\end{tabular} \\\hline
	\end{tabularx}
\end{table}

\subsubsection{Diagnosis results}
\begin{figure}[htb]
	\centering  
	\subfigure[]{
		\includegraphics[width=0.48\columnwidth]{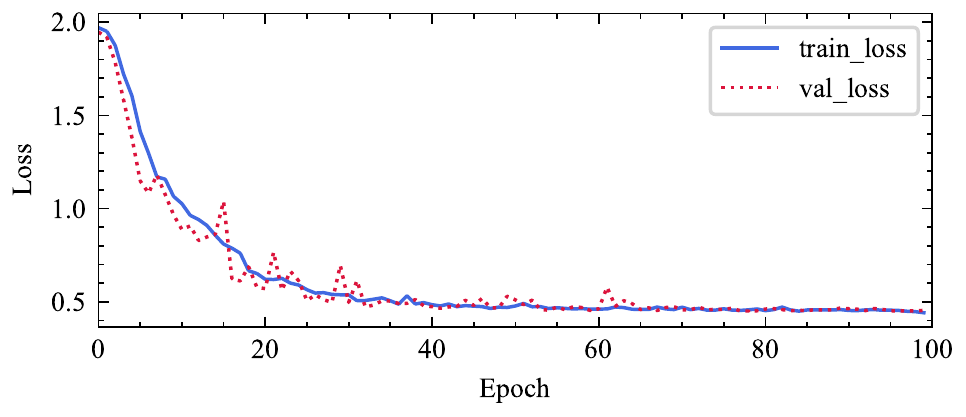}}
	\subfigure[]{
		\includegraphics[width=0.48\columnwidth]{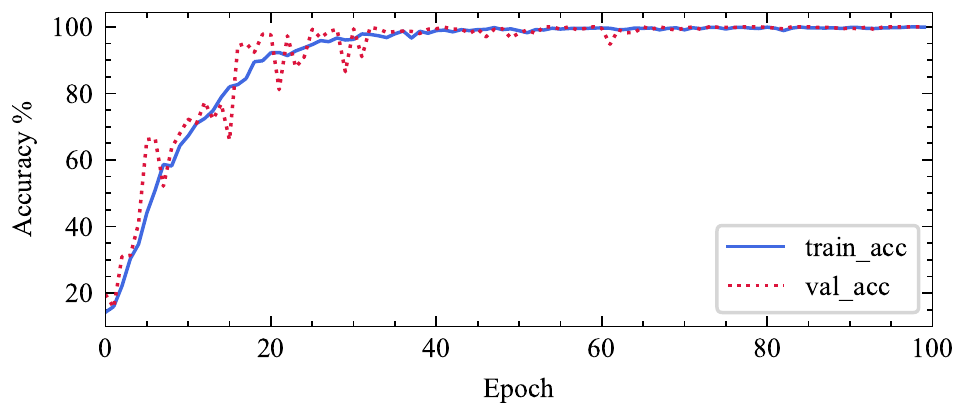}}
	\caption{The (a) loss and (b) accuracy curves of the proposed TFT.}
	\label{fig:lossHistory}
\end{figure}

Based on the optimal network structure and hyperparameter setting, the TFT model is trained on ABLT-1A Bearing Dataset 6308. As can be seen from Fig. \ref{fig:lossHistory}, after about 40 epochs, accuracy and loss values of the training and validation datasets have been very stable, and the model has started to converge, indicating the strong convergence ability of the TFT. Note that in the early stage of training, the training loss is larger than the validation loss because the use of dropout limits the model capacity in training. With the process of training, dropout will drive the network to learn more robust features. Finally, the training loss and validation loss of the network are basically stable at the same value, indicating the good generalization ability of the network. The regularization technique we used fully guarantees the robust generalization of the network.

Then, the trained model is used to classify the testing dataset to evaluate performance. Training and test have been repeated 10 times on TFT and three benchmark models. We draw the results in Table \ref{tab:performance6308}. Besides, the total number of trainable parameters and the average training time are also calculated to comprehensively compare the performance of the models. Comparing the test performance of the 4 models, the proposed TFT achieves the best prediction accuracy. Its maximum prediction accuracy can reach 100\%, and the average accuracy is also the highest. The accuracy variance of TFT is smaller, indicating that the prediction result is more stable. The performance of RNN is second only to TFT, whose maximum accuracy is 100\% and average accuracy is lower than that of TFT. Meanwhile, the variance of RNN is larger, so the result is not as stable and reliable as TFT. The prediction accuracy of DNN and CNN is significantly lower than that of TFT and RNN, which is obviously related to the characterization ability of the model itself. DNN and CNN are just simple multiplication or convolution operation on inputs, lacking the grasp of temporal information. 

\begin{table}[htb]
	\centering
	\caption{Test performance, size and training time usage of the models on ABLT-1A Bearing Dataset6308.}
	\label{tab:performance6308}
	\footnotesize
	\begin{tabularx}{\columnwidth}{p{1cm}<{\centering}p{2.5cm}p{2.5cm}p{2.5cm}p{2.5cm}l}
		\hline
		Model & Mean accuracy & Best accuracy & Std  & Params num & Training time (s) \\\hline
		TFT   & 99.94\%       & 100.00\%      & 0.05 & 335,016    & 690               \\
		DNN   & 80.15\%       & 85.71\%       & 3.81 & 25,757,191 & 740               \\
		CNN   & 92.56\%       & 97.83\%       & 0.55 & 11,176,839 & 1030              \\
		RNN   & 97.03\%       & 100.00\%      & 1.56 & 1,844,103  & 1800              \\\hline
	\end{tabularx}
\end{table}

Furthermore, we compare the scale and training time of these models. DNN contains the most trainable parameters because of its internal fully connected structure. This will lead to the model over parameterized, thus make it easier to overfit when the number of samples is limited. Interestingly, the parameters of CNN are slightly less than that of DNN, but the training time is much longer than that of DNN. This is obviously due to the fact that although convolutional structure reduces the number of trainable parameters through weight sharing, calculation amount corresponding to each parameter increases significantly. The number of trainable parameters in RNN is much less than that in CNN and DNN, but the training time of RNN is the longest for its non-parallel computation.

Generally speaking, the accuracy of TFT and RNN which can grasp the temporal information is higher. {But for RNN, this comes at the cost of a very long training time.} The results show that the proposed TFT method based on the Transformer structure can process the time series information well with higher prediction accuracy. Moreover, TFT can realize parallel computing, so its training is much faster than models with recurrent structure.

\begin{figure}[htb]
	\centering 
	\subfigcapskip=-10pt %设置子图与子标题之间的距离 
	\subfigure[]{
		\includegraphics[width=0.4\columnwidth]{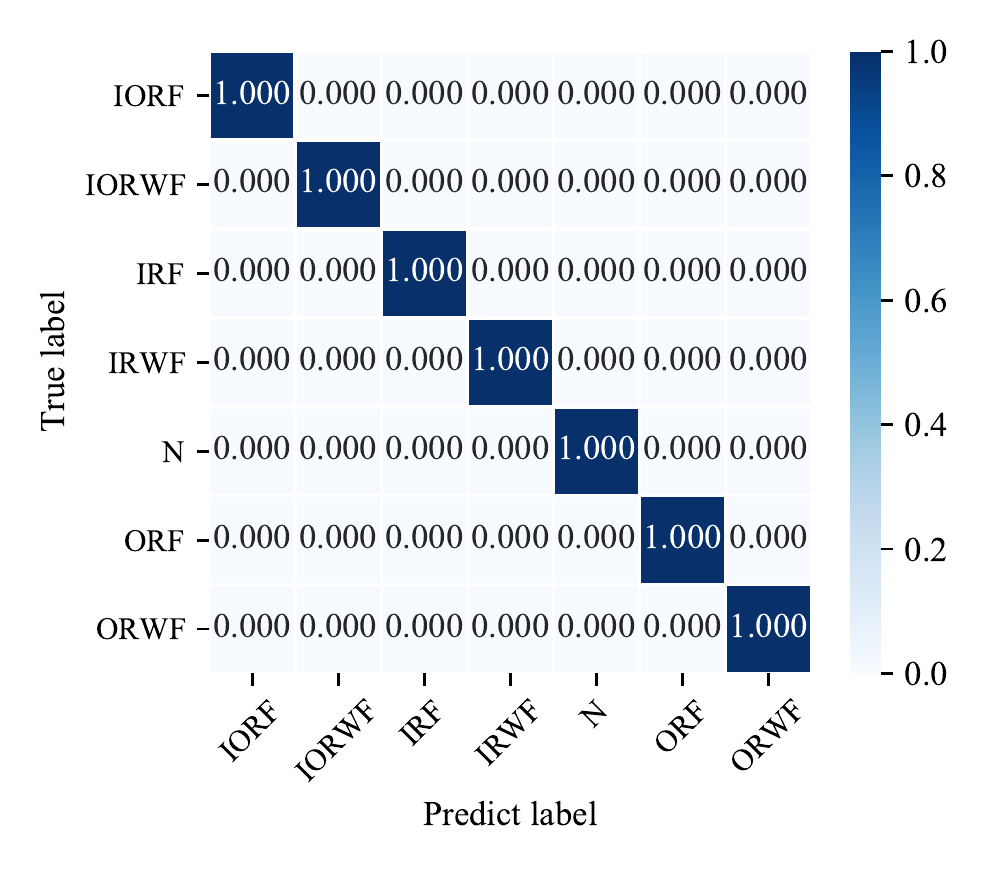}}
	\subfigure[]{
		\includegraphics[width=0.4\columnwidth]{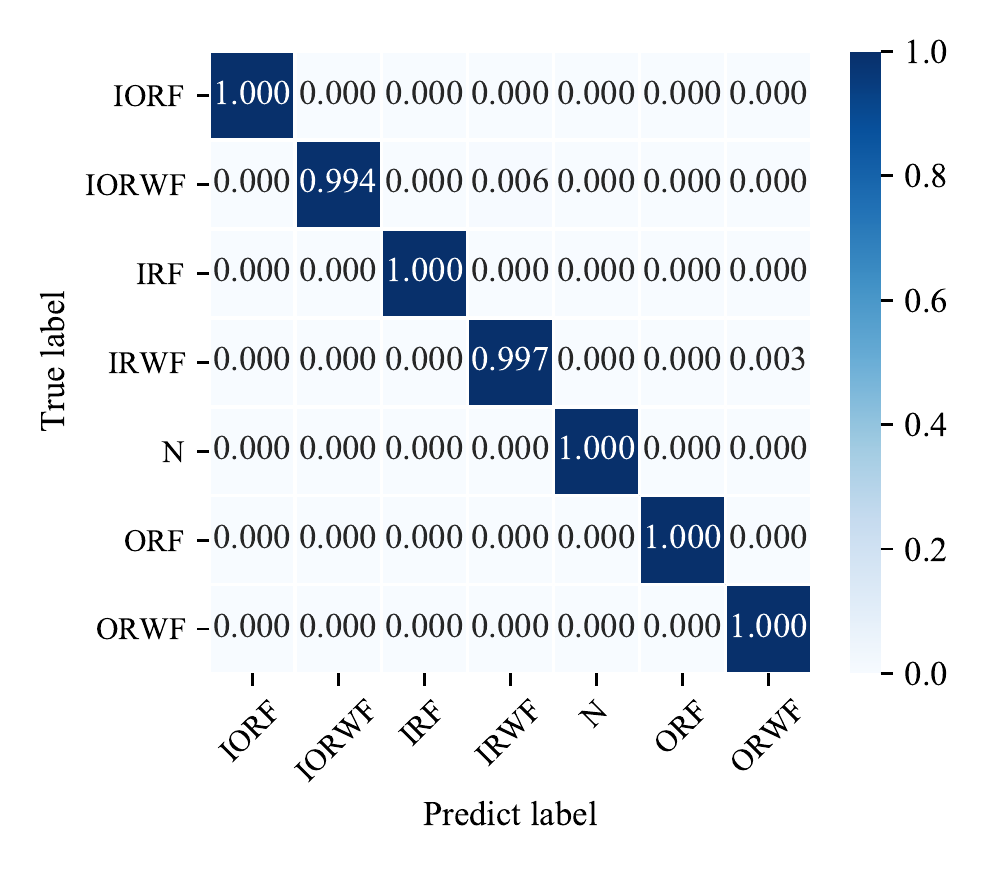}}
	\caption{Confusion matrix of (a) the best and (b) the worst results of time--frequency Transformer on ABLT-1A Bearing Dataset 6308.}
	\label{fig:confusion6308}
\end{figure}

Fig. \ref{fig:confusion6308} shows the confusion matrix of the best and the worst test results of TFT, with the accuracy of 100\% and 99.88\%, respectively. The columns represent the predict labels while the rows represent the true labels for different health states. In the best case, the accuracy is 100\% for each health condition. In the worst one, the accuracy is 100\% for most conditions except for IORWF and IRWF. A small number of samples with IORWF and IRWF were misjudged as IRWF and ORWF, respectively, {indicating greater challenge of weak fault classification}. It is also worth noting that, even in the worst case, TFT does not fail to identify the three non-weak fault states and the normal state.

\begin{figure}[htb]
	\centering 
	%	\subfigcapskip=-10pt %设置子图与子标题之间的距离 
	\subfigure{
		\includegraphics[width=0.9\columnwidth]{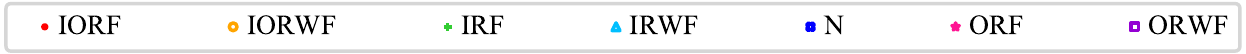}} \\
	\setcounter{subfigure}{0}
	\subfigure[]{
		\includegraphics[width=0.23\columnwidth]{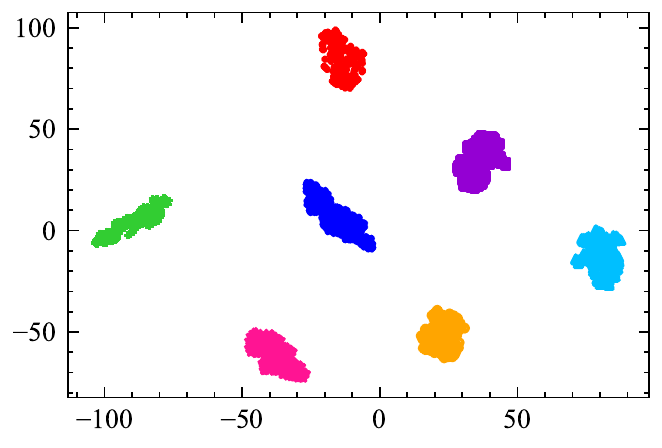}}
	\subfigure[]{
		\includegraphics[width=0.23\columnwidth]{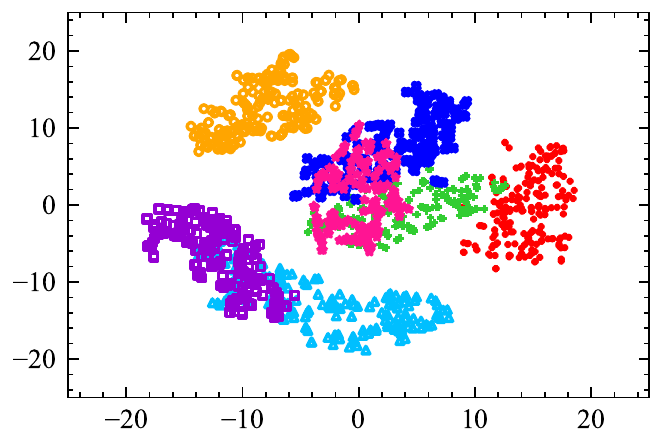}}
	\subfigure[]{
		\includegraphics[width=0.23\columnwidth]{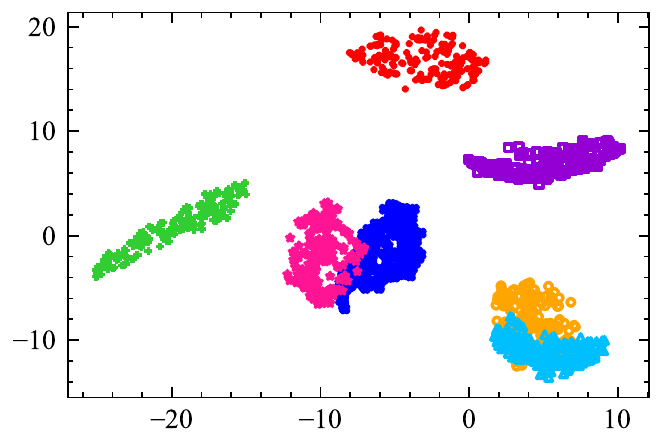}}
	\subfigure[]{
		\includegraphics[width=0.23\columnwidth]{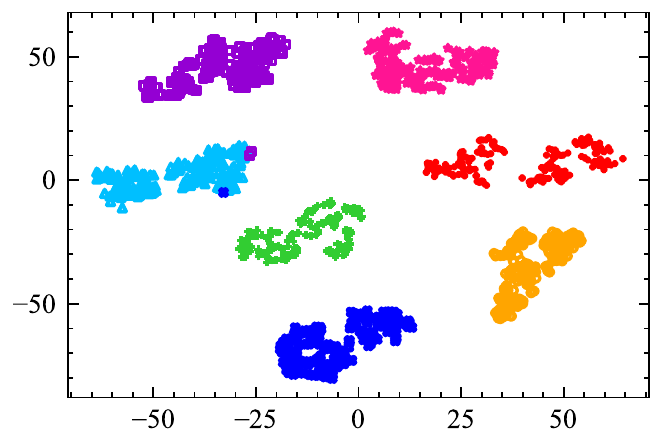}}
	\caption{The 2D visualization result of learned features via (a) TFT (b) DNN (c) CNN (d) RNN.}
	\label{fig:tsne6308}
\end{figure}

To visualize the extracted features of these models, t-distributed stochastic neighbor embedding (t-SNE) \cite{vandermaaten2008} is used to simplify the extracted high-dimensional features of the last hidden layer extracted by the above four methods into two-dimensional vector distribution, the visualization results of the test samples are shown in Fig. \ref{fig:tsne6308}. As indicated that only the hidden features extracted by the TFT algorithm can accurately separate all faults.

\subsubsection{Visualization of attention weights}
\begin{figure}[htb]
	\centering  
	\subfigtopskip=2pt %设置子图与上面正文或别的内容的距离
	\subfigbottomskip=0pt %设置第二行子图与第一行子图的距离，即下面的头与上面的脚的距离
	\subfigcapskip=0pt %设置子图与子标题之间的距离
	\subfigure[]{
		\label{fig:attn.a}
		\includegraphics[width=0.2\columnwidth]{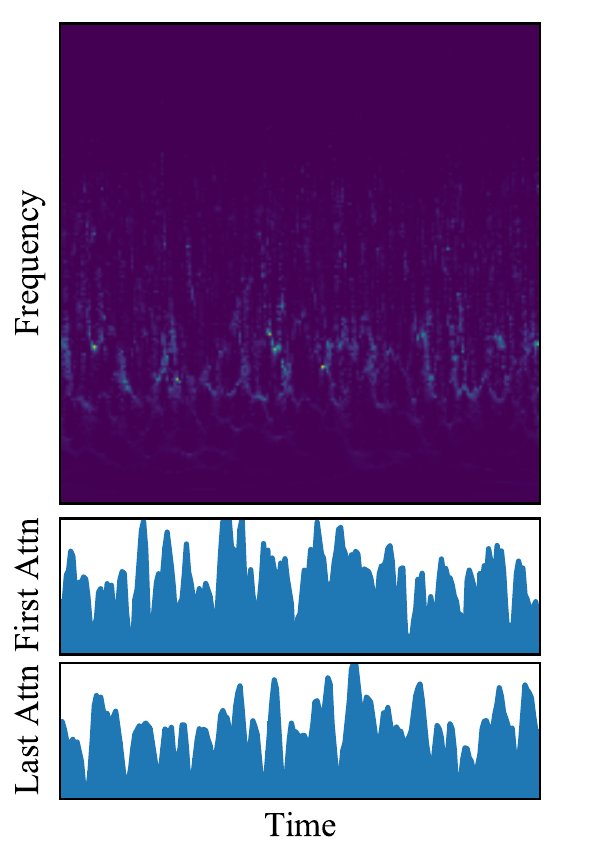}}
	\subfigure[]{
		\label{fig:attn.b}
		\includegraphics[width=0.2\columnwidth]{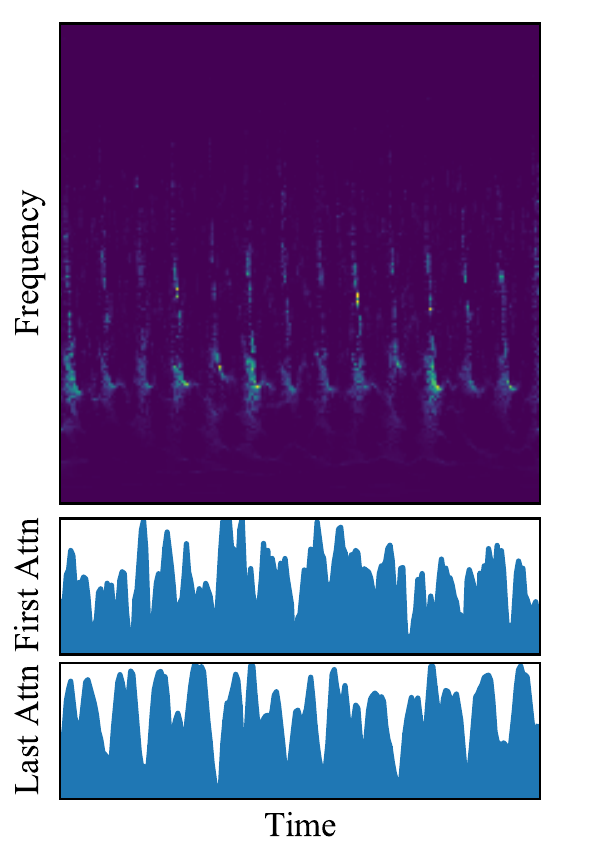}}
	\subfigure[]{
		\label{fig:attn.c}
		\includegraphics[width=0.2\columnwidth]{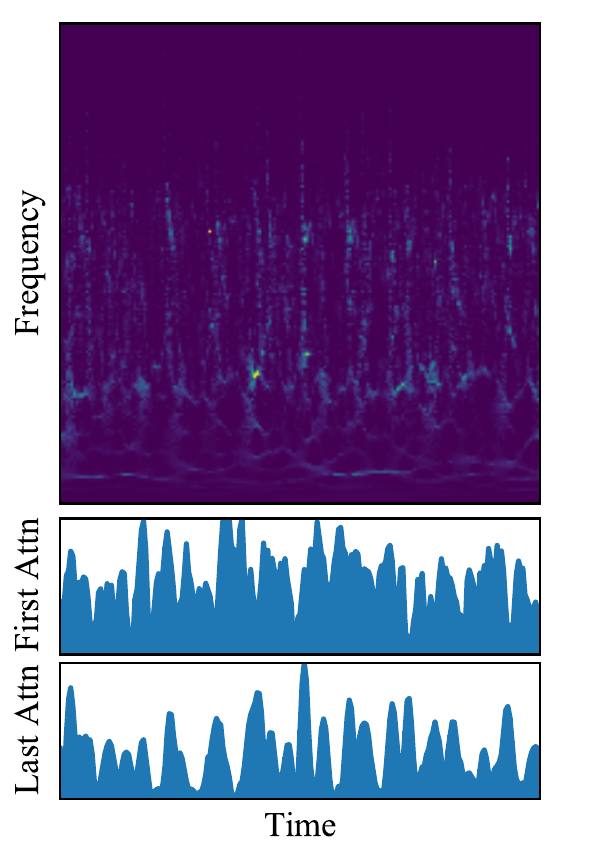}}
	\subfigure[]{
		\label{fig:attn.d}
		\includegraphics[width=0.2\columnwidth]{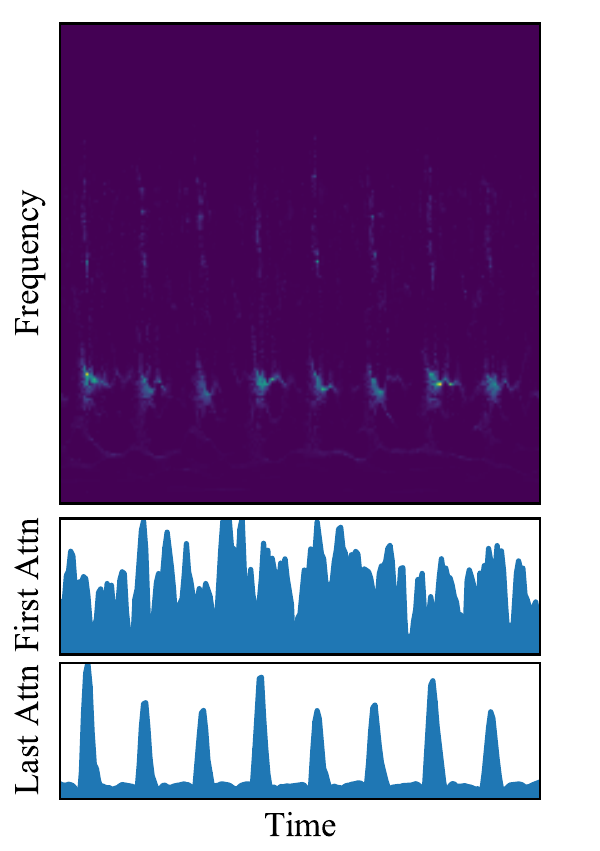}}
	\subfigure[]{
		\label{fig:attn.e}
		\includegraphics[width=0.2\columnwidth]{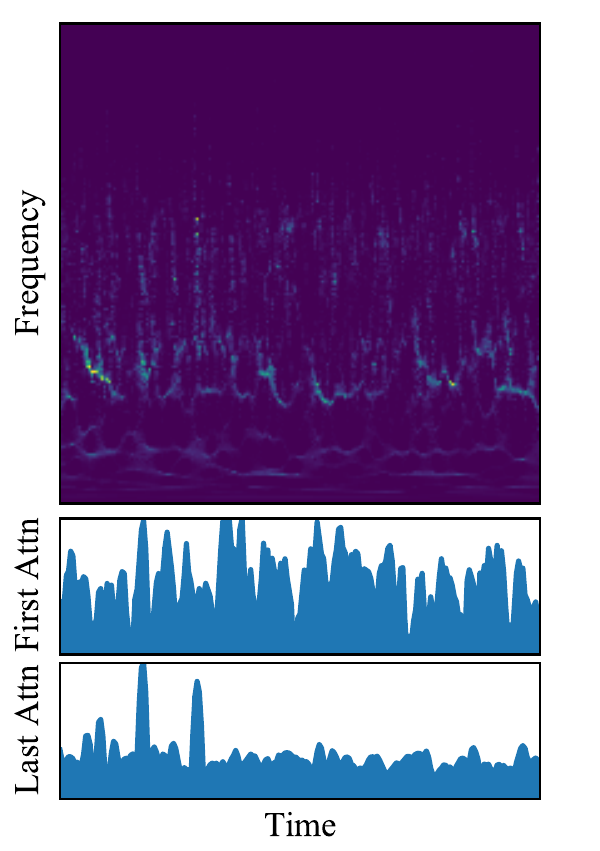}}
	\subfigure[]{
		\label{fig:attn.f}
		\includegraphics[width=0.2\columnwidth]{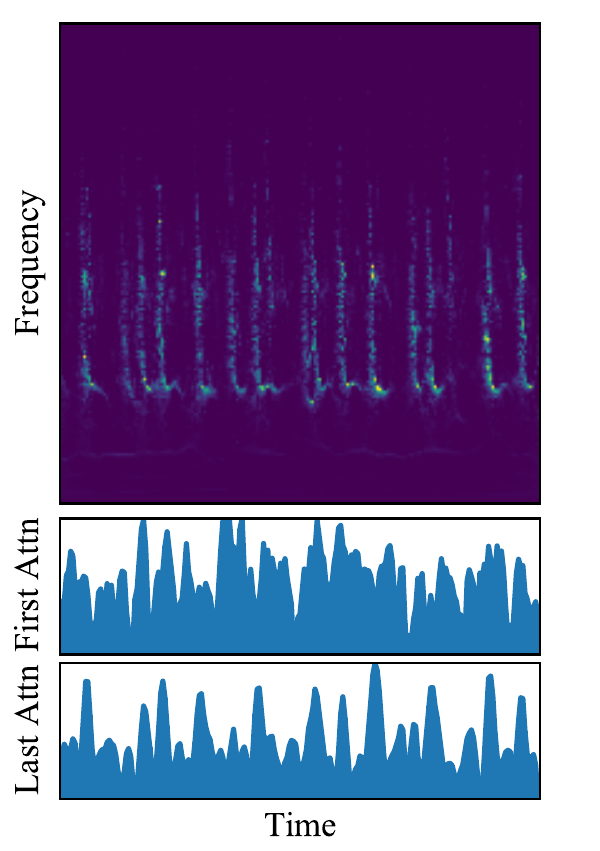}}
	\subfigure[]{
		\label{fig:attn.g}
		\includegraphics[width=0.2\columnwidth]{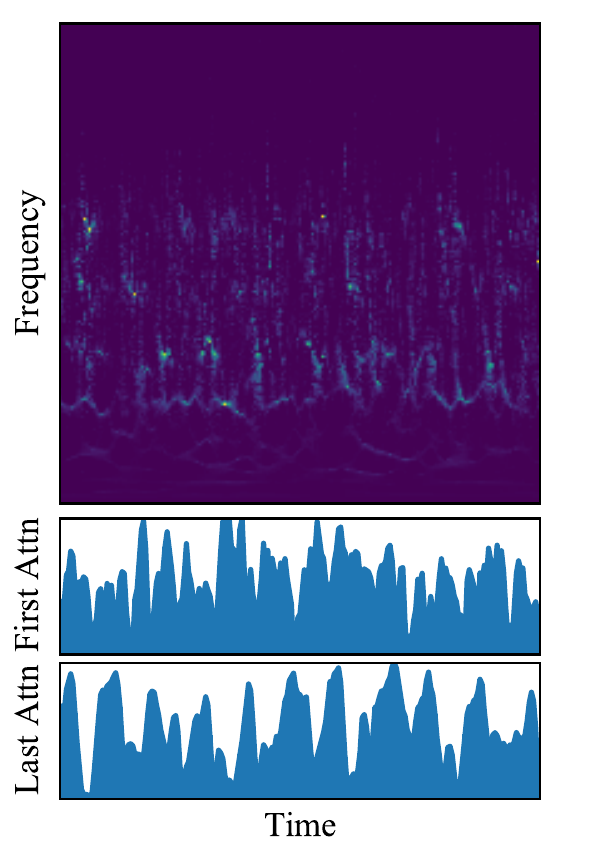}}
	\caption{{Attention weights visualization of bearings (a) 6308N (b) 6308IRF (c) 6308IRWF (d) 6308ORF (e) 6308ORWF (f) 6308IORF (g) 6308IORWF.}}
	\label{fig:attentions}
\end{figure}

Since the proposed TFT extracts features from TFRs completely based on attention mechanism, it is necessary to explore its mechanism by attention visualization. The attention mechanism adopted by Transformer is mainly used to exert different degrees of attention on the tokens and form the relationship between different tokens. Here, we try to show the attention degree of attention mechanism on different tokens, namely attention weight. Firstly, the attention weight tensors of the first and last self--attention layers are derived. Note that the attention weight is not the output of the attention layer, but the weight ${\rm softmax}\left(Q_s K_s^{\top}/\sqrt{d_k} \right) $ of the input in Eq. \ref{eq:mha}. Since the calculation results of multi-head attention are concatenated in the network, we sum the weight matrix of $h$ attention heads. Furthermore, the target of attention heads is tokens, and our TFT uses time--frequency cutting patches as tokens. Therefore, the attention mechanism actually works along the time direction. The attention mechanism and the TFRs of bearing signals are drawn in Fig. \ref{fig:attentions}. These figures show the normalized attention weight of the first and last attention layers on different tokens. The larger the value is, the greater the attention weight is.

It can be seen from Fig. \ref{fig:attentions} that the attention weight distribution of different fault samples in the first transformer block is almost the same. This obviously adheres to our intuition since the network cannot distinguish different fault types in the input layer, but “observe” different samples with the same strategy. With the layer-by-layer attention processing, the network will be able to attach different attention weights to different fault types. Combined with the TFRs, it can be found that the last Transformer block focuses on the tokens with larger values, that is, it pays more attention to the time when the amplitude is more obvious. Through such concentration of attention, the TFT model can effectively grasp the characteristic information from the TFRs, and observe each token with different weights. Thus, TFT can accurately extract the key features of different fault types and avoid the interference of fault independent factors.

\subsubsection{Anti-noise robustness test}
\begin{figure}[htb]
	\centerline{\includegraphics[width=0.6\columnwidth]{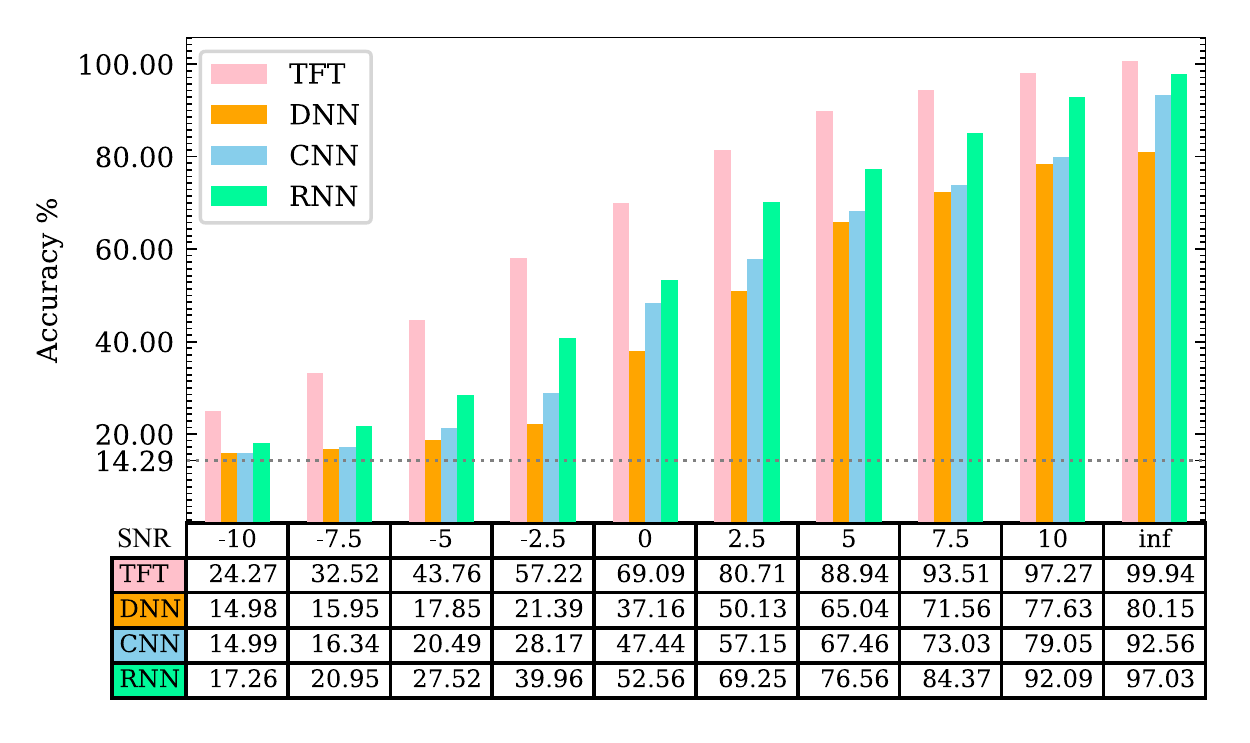}}
	\caption{Average diagnostic accuracy for different fault diagnosis methods under different SNRs.}
	\label{fig:snrs}
\end{figure}

To evaluate the anti-noise performance and robustness of the designed algorithm and fault diagnosis method, we operate an anti-noise robustness test. Specially, different degrees of signal-to-noise (SNR) is added into the original bearing signals. The same experimental setup was used to evaluate the variation in performance of each fault diagnosis methods above with different SNR Settings. By adding different degrees of noise to the vibration signal, this experiment evaluates the ability of these fault diagnosis methods to adapt to a noise-changing environment. As shown in Fig. \ref{fig:snrs}, it is worth noting that the proposed method achieves the best performance. When SNR is bigger than 5, that is, under the condition of relatively small noise, the proposed TFT can provide an acceptable prediction accuracy. With the SNR decreasing, the accuracy of all these methods decreases. When the SNR is smaller than -5, the accuracy of DNN and CNN almost decreased to a meaningless value. Note that the meaningless value means the accuracy almost equal to 14.29\% of guess blindly. To sum up, the proposed TFT method has relatively high diagnosis accuracy under noise interference environment, and its accuracy is higher than those of other benchmark methods. Also, the proposed TFT method is less disturbed by noise, and the accuracy decreases more slowly with the increase of noise.

\subsection{ABLT-1A Bearing Dataset 6205}
\subsubsection{Dataset and benchmark test description}
The rolling bearing HRB6205 is selected as the experimental bearing in this experiment. {Different the previous case, a triaxial acceleration sensor with the sampling frequency of 12,000Hz is used to collect vibration signals of fault or normal rolling bearing.} Thus, the original vibration signal is converted into three-channel digital signal by a data acquisition card. {In addition, we have taken into account the different rotational speeds of 1200 rpm, 1500 rpm, 1750 rpm, 2000 rpm.} The specific experimental dataset is described as follows. Under zero-load conditions, six types of fault conditions such as normal (6205N), inner ring fault (6205IRF), outer ring fault (6205ORF), ball fault (6205BF), inner and outer ring compound fault (6205IORF), and outer ring and ball compound fault (6205ORBF) are simulated respectively. We obtained 1000 samples with a length of 1024 for each health state and each speed. That is, a total of $1000 \times 6 \times 4 = 24000$ samples. To distinguish the samples under different speeds, the dataset is divided into four subsets according to the speed, and each subset contains 6000 samples. The benchmark will be conducted on each sub-dataset. 60\% of each subset set is used as training dataset, 20\% as validation dataset, and 20\% as test dataset for final test. In each training and testing, the subsets are randomly divided to ensure the comprehensive evaluation of the model performance.

The proposed TFT model and three benchmark methods in Case 1 are used in the benchmark test. The model structure and hyperparameter setting are basically the same as the optimal setting in Case 1. Similarly, SWT is used to process the vibration signals of each sample to obtain the TFRs. Differently, it should be noted that the signals from the three channels are processed separately and then stacked into multi-channel TFRs to make full use of the multi-channel information in the dataset. After downsampling by bicubic interpolation algorithm, the input data with a shape of $224 \times 224 \times 3$ is obtained. The proposed TFR and CNN can directly process multi-channel input, while DNN and RNN need flattening operation in the input layer.

\subsubsection{Diagnosis with multi-channel TFRs}
Based on the optimal network structure and hyperparameter setting, the TFT model is trained respectively on four subsets of Bearing Dataset 6205. In addition, the three benchmark models are fully trained for comparison. The fully trained models are applied to the diagnosis of test set samples in each subset, and the average results are shown in Table \ref{tab:performance6205}. Note that the test for each subset is repeated five times, and the table shows the average accuracy of all subsets.

It can be seen that the accuracy of TFT is higher when multi-channel data is used. Both TFT and CNN can directly input multi-channel data and construct feature graphs fusing multi-channel information. When DNN and RNN are dealing with larger dimension input data, the large network size obviously limits the generalization ability. Besides, the large amount of multi-channel input data does not significantly increase the training time of TFT and CNN due to the parallel computing capability. Particularly, hardly has the training time of TFT increased. Since RNN has no parallel processing ability and no learnable embedding module, its training time is greatly increased. It can be seen that compared with other benchmark models, the proposed TFT can better deal with multi-channel TFRs and obtain better diagnosis performance with lower computational cost and network size.

\begin{table}[htb]
	\centering
	\caption{Test performance, size and training time usage of the models on ABLT-1A Bearing Dataset 6205.}
	\label{tab:performance6205}
	\footnotesize
	\begin{tabularx}{\columnwidth}{p{2cm}<{\centering}p{4.5cm}p{4.5cm}l}
		\hline
		& Average & Params num & Training time (s) \\\hline
		TFT & 99.97\% & 363,431    & 750               \\
		DNN & 67.17\% & 77,137,286 & 1,510             \\
		CNN & 92.65\% & 11,182,598 & 1,320             \\
		RNN & 96.38\% & 16,368,134 & 3,700             \\\hline
	\end{tabularx}
\end{table}

\subsubsection{Diagnosis across different conditions}
\begin{figure}[htb]
	\centerline{\includegraphics[width=0.4\columnwidth]{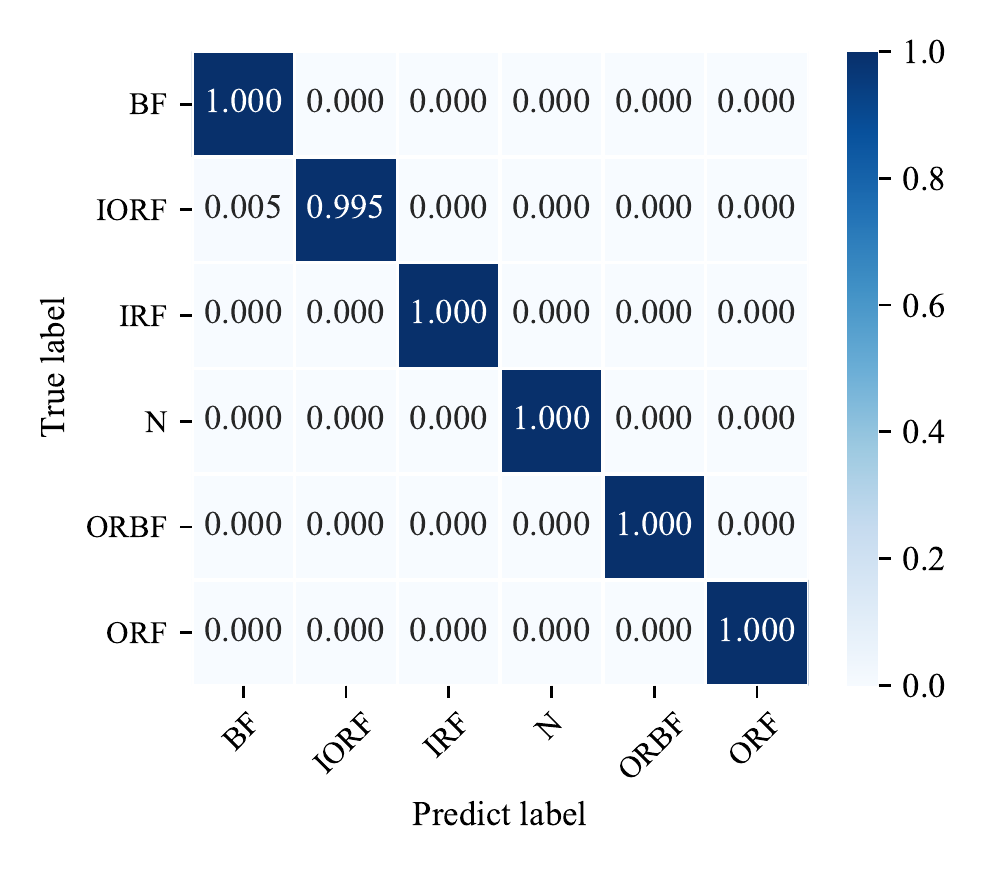}}
	\caption{Confusion matrix of the best results of time--frequency Transformer on multiple speed ABLT-1A Bearing Dataset 6308.}
	\label{fig:confusion6205}
\end{figure}

Bearing Dataset 6205 contains the vibration signals of bearings collected at different speeds, which enables us to test the performance of TFT diagnosis under multiple speed conditions. Different from the previous benchmark test, which divided the dataset into four subsets, we mixed and scrambled all samples with different speeds in this test. Therefore, the mixed dataset contains 24000 samples in total. Based on the optimal TFT structure and hyperparameter setting, the model training is repeated 10 times and used for inference of test samples. The final result shows that the average classification accuracy is 99.87\%, and the highest accuracy is 99.92\%. The confusion matrix of the best results is shown in Fig. \ref{fig:confusion6205}. Only a few samples with the real label of IORF were misclassified into BF.

\begin{figure}[htb]
	\centerline{\includegraphics[width=0.5\columnwidth]{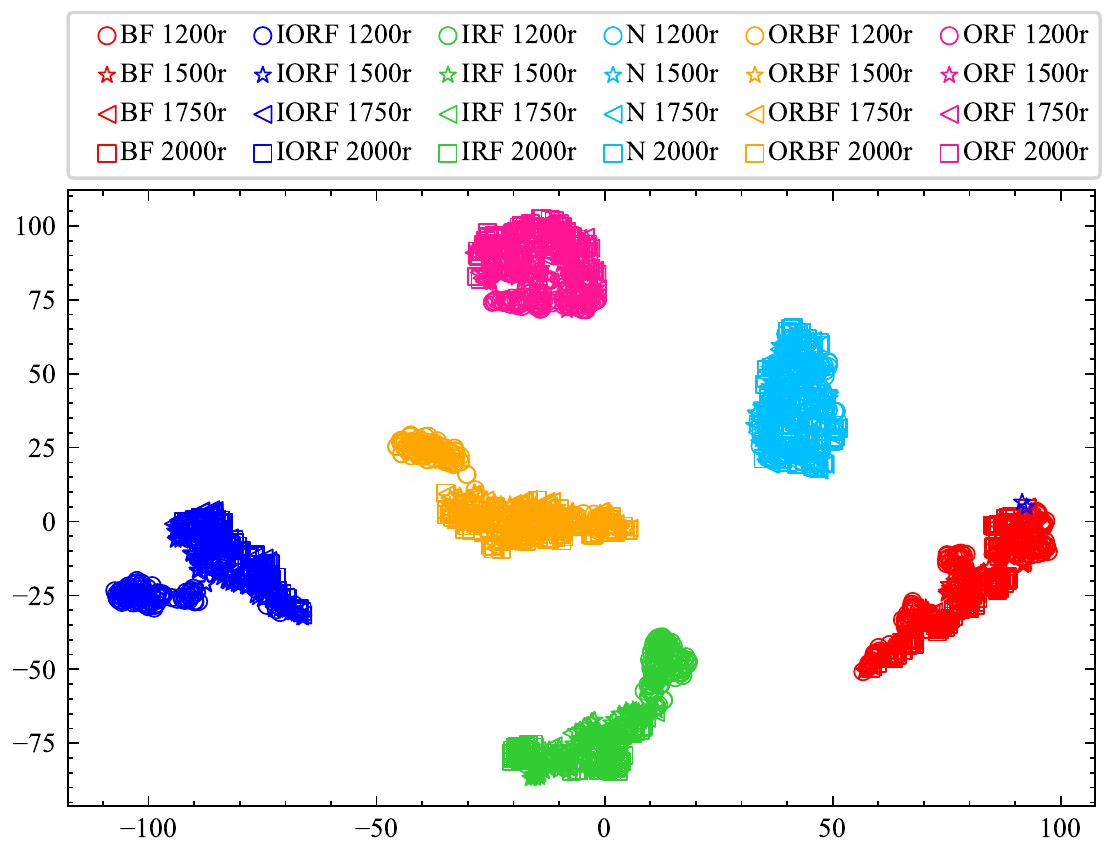}}
	\caption{The 2D visualization result of learned features from different rotate speed.}
	\label{fig:tsne6205}
\end{figure}

To further investigate the fault feature extraction of TFT under multiple speed conditions, t-SNE is used to visualize the hidden features as shown in Fig. \ref{fig:tsne6205}. Different colors in the figure represent different health states, while different markers represent different speeds. It can be seen that the hidden features after highly abstracted with TFT have good distinguishability. The samples of different health states can be well distinguished. Furthermore, the samples of the same state, even if with different speeds, can concentrate well. Only in the samples with IORF and ORBF faults, the 1200r samples are slightly separated from the samples with other speeds, but this does not prevent the classifier from classifying them into one category. In addition, several IORF 1500r samples that were misclassified as BF failures can also be clearly observed.

\subsection{Comparison with state-of-the-art}
{The above comparisons with three representative benchmark models significantly demonstrates the advantages of TFT in terms of feature extraction capability, network size, training speed, classification accuracy, and robustness. These comparisons are mainly aimed at the performance play of the network itself, while may be not sufficient for specific target tasks. Therefore, comparison with other fault diagnosis methods in practical application is necessary to verify the effectiveness of TFT-based diagnosis.}

{In this section, several state-of-the-art techniques for fault diagnosis are employed to comparison on Dataset6308 and Dataset 6205.} These comparison methods include Li’s extreme learning machine (ELM) \cite{li2016}, Zhang’s support vector machine (SVM) \cite{zhang2015}, Xu’s deep convolutional neural networks (DCNN) \cite{xu2020}, Chen's convolutional neural networks (CNN) \cite{chen2019a}, Shao’s deep belief network (DBN) \cite{shao2015}, Chen’s long short-time memory (LSTM) network \cite{chen2017}, Liu's GRU-based non-linear predictive denoising autoencoder (GRU-NP-DAE) \cite{liu2018}, Shao’s auto-encoder (AE) \cite{shao2017} and Mao’s auto-encoder extreme learning machine (AE-ELM) \cite{mao2017}. {Most of these methods are based on the latest CNN and RNN type models and utilize different inputs, e.g. vibration signals, time domain statistics, frequency domain, and time--frequency domain features, respectively.}

\begin{table}[]
	\centering
	\caption{Performance of the comparison methods on the two datasets.}
	\label{tab:sota}
	\footnotesize
	\begin{tabularx}{\columnwidth}{Xp{5cm}p{2cm}<{\centering}p{2cm}<{\centering}p{2cm}<{\centering}}
		\hline
		Diagnose methods 		& Signal processing                & Dataset6308      & Dataset6205      & Average          \\\hline
		ELM \cite{li2016}		& MRSVD   (multi-resolution SVD)   & 79.58\%          & 71.95\%          & 75.77\%          \\
		SVM \cite{zhang2015}	& EEMD (ensemble   EMD)            & 88.90\%          & 87.94\%          & 88.42\%          \\
		DCNN \cite{xu2020}		& VMD                              & 93.67\%          & 95.82\%          & 94.75\%          \\
		{CNN \cite{chen2019a}}	& DWT (discrete wavelet transform) & 97.54\%		  & 98.16\%			 & 97.85\%			\\
		DBN \cite{shao2015}		& Time-domain statistic features   & 98.20\%          & 99.75\%          & 98.98\%          \\
		LSTM \cite{chen2017}	& EMD                              & 99.12\%          & 97.54\%          & 98.33\%          \\
		AE-ELM \cite{mao2017} 	& FFT                              & 91.03\%          & 87.58\%          & 89.31\%          \\
		AE \cite{shao2017}		& Vibration signals                & 93.59\%          & 92.81\%          & 93.20\%          \\
		{GRU-NP-DAE \cite{liu2018}} & Vibration signals			   & 96.17\%		  & 95.38\%		     & 95.78\%			\\\hline
		TFT              & SWT                              & \textbf{99.94\%} & \textbf{99.97\%} & \textbf{99.96\%} \\\hline
	\end{tabularx}
\end{table}

These state-of-the-art methods are used for Dataset 6308 and Dataset 6205, respectively, whose results are shown in Table \ref{tab:sota}. {The performance of these methods varies using different data processing methods and diagnostic models. Among them, models with the ability to grasp temporal information, such as RNNs and their variants, clearly bring more significant performance gains compared to traditional ELMs and SVMs. Besides, TFT outperforms other DL-based approaches even all receive time--frequency inputs, which is consistent with the benchmark in case studies before.} Among all these state-of-the-art solutions, the proposed TFT achieves the highest classification accuracy on both datasets, which further proves the superiority of the proposed method on rolling bearing fault diagnosis.

\section{Conclusion}
In this paper, we proposed a new time--frequency Transformer (TFT) inspired by vanilla Transformer architecture to process time--frequency representations (TFRs). Based on TFT, this paper presented an end-to-end fault diagnosis framework for rolling bearings. In this method, the vibration signals of rolling bearing are processed by SWT to obtain multi-channel TFRs, and then the TFRs are input into TFT to extract discriminative hidden features and accurately classify fault modes.

The proposed TFT has the following characteristics: 1) TFT completely abandons the tedious recurrence structure and convolution operation commonly used in traditional DL-based frameworks, completely relying on multi-head self--attention mechanism and feed-forward neural network layers. Compared with classical CNN and RNN, this greatly improves the parallel computing ability of the network and reduces the network scale. 2) The feature extraction structure based on self--attention mechanism with residual connector can accurately focus on the effective feature areas in the input graphs. Compared with RNNs that with long-range dependent defects, this model can better establish the relationship in input sequences.

The effectiveness of the proposed fault diagnosis method of rolling bearing based on TFT is verified by case studies on several experimental data in this paper. Compared with other benchmark models and state-of-the-art methods, the superiority of this method is verified. The diagnosis framework has the following advantages: 1) Compared with other methods based on classical deep learning model, this method has higher diagnosis accuracy and faster training speed. 2) This method can make better use of the collected multi-channel signals, and contribute this advantage to higher diagnostic accuracy and efficiency. 3) This method can adapt to a certain degree of noise environment and provide effective fault diagnosis under strong noise conditions. 4) This method can be used for fault diagnosis under multiple working conditions (speeds).

Considering that there is still much room for improvement, our future work will focus on the following. 1) We will try to apply Transformer architecture to the prognostic field. 2) We will try to further improve the model, such as using convolution operation in the tokenizer module to improve the local receptive field.

\section*{Acknowledge}
The authors gratefully acknowledge the financial support of the National Natural Science Foundation of China (No. 52075095).

\bibliographystyle{elsarticle-num.bst}

\bibliography{bibliography}

\end{document}